\documentclass{article} 
\usepackage{iclr2027_conference,times}

\usepackage{amsmath,amsfonts,bm}

\def\eqref#1{equation~\ref{#1}}

\def\1{\bm{1}}

\DeclareMathAlphabet{\mathsfit}{\encodingdefault}{\sfdefault}{m}{sl}
\SetMathAlphabet{\mathsfit}{bold}{\encodingdefault}{\sfdefault}{bx}{n}

\usepackage{hyperref}
\usepackage{url}
\usepackage{xcolor}
\usepackage{colortbl}
\usepackage{soul}
\usepackage{graphicx}
\usepackage{booktabs}
\usepackage{multirow}
\usepackage{wrapfig}
\usepackage{amsmath}
\usepackage{amssymb}
\usepackage{silence}
\usepackage{microtype}
\usepackage{placeins}
\usepackage{enumitem}
\usepackage{CJKutf8}
\usepackage{fontawesome5}
\usepackage{tcolorbox}
\tcbuselibrary{breakable,listings,skins}

\newtcblisting{promptbox}[1]{%
  listing only, colback=white, colframe=black,
  colbacktitle=white, coltitle=black,
  title={\small\bfseries #1}, fonttitle=\small\bfseries, titlerule=0.4pt,
  listing options={basicstyle=\ttfamily\scriptsize, breaklines=true,
    breakautoindent=false, breakindent=0pt, breakatwhitespace=true,
    postbreak=\mbox{$\hookrightarrow$\space},
    columns=fullflexible, keepspaces=true,
    extendedchars=true, inputencoding=utf8,
    literate={—}{{---}}3 {–}{{--}}2 {‘}{{`}}1 {’}{{'}}1 {“}{{``}}2 {”}{{''}}2 {…}{{...}}3 {→}{{$\rightarrow$}}1,
    aboveskip=0pt, belowskip=0pt},
  boxrule=0.4pt, arc=0pt, left=6pt, right=6pt, top=2pt, bottom=2pt, breakable,
}

\newif\ifshowtbd
\showtbdtrue

\title{Does Learning to Predict the World Help Agents Act? Auditing
World-Model Post-Training}

\author{
  Xinyu Che\textsuperscript{*,1} \quad
  Hang Yan\textsuperscript{*,1} \quad
  Yanchen Liu\textsuperscript{*,2} \\
  \textbf{
  Haochen Liu\textsuperscript{3} \quad
  Ruifeng Li\textsuperscript{2} \quad
  Anran Shi\textsuperscript{4} \quad
  Heng Wang\textsuperscript{1} \quad
  Jun Liu\textsuperscript{\dag,1}} \\
  \textsuperscript{1}Xi'an Jiaotong University \\
  \textsuperscript{2}University of Southern California \\
  \textsuperscript{3}University of the Chinese Academy of Sciences \\
  \textsuperscript{4}East China Normal University \\
  \texttt{\{xinyuche,hyan\}@stu.xjtu.edu.cn}
}

\iclrfinalcopy 
\begin{document}
\raggedbottom

\maketitle
\lhead{Preprint}
\begingroup
\renewcommand{\thefootnote}{}
\footnotetext{*~Equal Contribution. ~~$^\dagger$~Corresponding Author.\\
  \hspace*{1.8em}\faGithub~\url{https://github.com/KosmoCHE/WM-PostTraining-Audit.git}}
\endgroup

\begin{abstract}
Predicting how an environment will change before acting is a natural route to better decision making for agents. Recent post-training methods therefore require agents to predict the next observation and turn that prediction into a reward or a direct supervision signal, which is called world model. Existing next-observation training methods help the agent to learn the environmental content. However, they additionally involve an optimization process, which may introduce several effects other than learning to predict the world. Consequently, \textbf{where the performance gain comes from during the training process remains an open question}. We answer this research question through  replacing true next-observation targets with in-distribution mismatched observations during the training process.  Across two interactive text environments, mismatched targets lower prediction accuracy by 15.3--61.6\% relative to ground-truth targets, yet retain substantial task gains over the base model. Compared with the base model, trained models consider more candidate actions and exhibit less looping. We also introduce a setting that replaces prediction-based rewards with independent random signals. This training expands task coverage (pass@64) even when the reward carries no environment information. We also generalize this finding to VisualWebArena, where random-reward training raises pass@64 by 14.3\% relative to the base model, without observation-matching rewards or an external multimodal teacher for reward construction.
\end{abstract}

\section{Introduction}
\label{sec:introduction}

Decision making \citep{hafner2020dreamer,schrittwieser2020muzero,hao2023rap} requires an agent to anticipate how its actions will change the environment. Once deployed as an agent in a persistent interactive environment, it must answer not only ``what text comes next?'' but also ``what will happen after this action?'' \citep{rwml,vagen}. However, standard next-token pretraining \citep{brown2020gpt3,llama31,qwen3} only teaches a language model to predict how text continues, without directly supervising action-conditioned state transitions.

Recent work makes this predictive ability an explicit objective in agent post-training. Reward-based methods use the agreement between predicted and observed states as a learning signal \citep{rwml,vagen}. Auxiliary objectives add state prediction, inverse dynamics, or transition prediction to policy optimization \citep{envrl,tapo,paw}. Direct supervision and self-distillation train models on prediction targets from collected interaction data \citep{ui_oceanus,spa_selfplay,comap}. Although their training procedures differ, they share the same expectation. An agent that predicts environmental changes more accurately should act more effectively.

However, improved task performance alone does not establish that more accurate world prediction caused the improvement. Comparisons before and after world-model post-training measure the combined effects of learning from correct prediction targets and additional training \citep{rwml,comap}. We therefore ask whether the task gains depend on correct environment information or can also arise from other effects of training.

To answer this question, we conduct controlled RL experiments on ALFWorld and ScienceWorld. If accurate world prediction drives better action, corrupting the prediction target should weaken both prediction accuracy and task performance. We evaluate prediction accuracy to check whether target corruption affects the agent's predictions. We then measure single-attempt success (pass@1) and task coverage within 64 attempts (pass@64).

\begin{figure}[!t]
  \begin{center}
  \includegraphics[width=\linewidth]{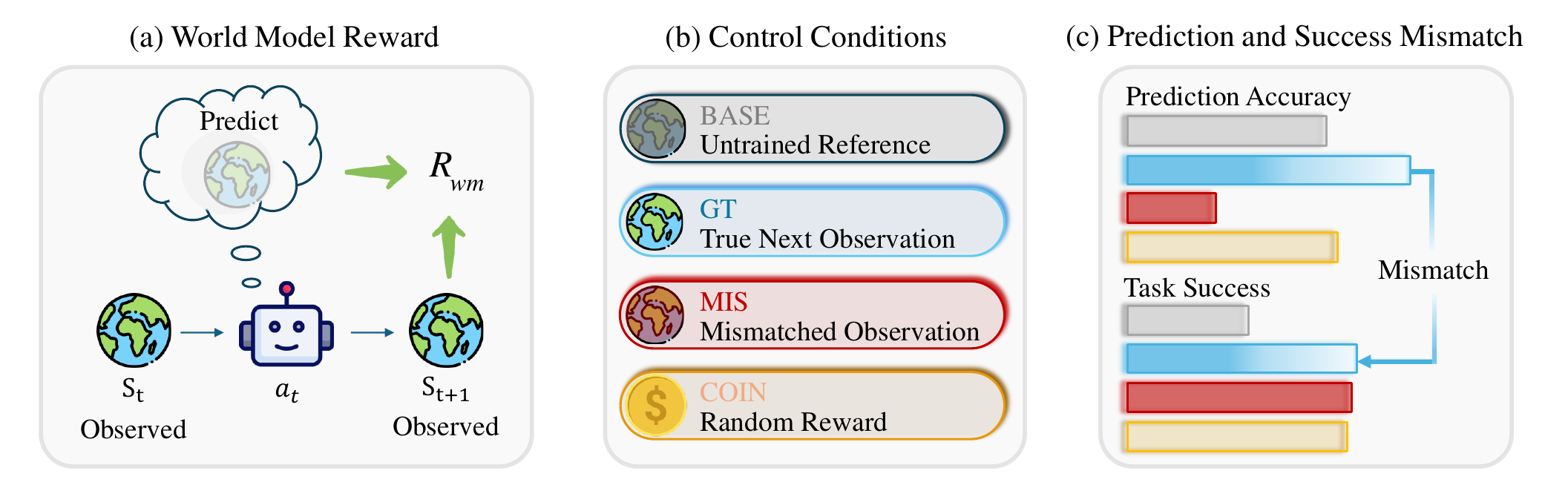}
  \end{center}
  \caption{Overview of the experimental design and main finding. \textbf{(a)} World-model post-training uses agreement between predicted and observed next states as a reward. \textbf{(b)} We conduct our analysis experiments in four conditions that isolate the role of correct prediction targets. \textbf{(c)} In ALFWorld, higher prediction accuracy does not necessarily imply better task performance. Agents trained with mismatched targets retain substantial task gains despite less accurate predictions.}
  \label{fig:audit_to_application}
\end{figure}

BASE provides a reference without post-training, while the ground-truth condition (GT) trains on true next observations. Their comparison measures the overall training gain. The mismatched condition (MIS) replaces the true targets with in-distribution mismatched observations under the same training procedure. These mismatched targets remain real observations from the same training distribution, but belong to different transitions,  breaking the correspondence between each input and its observed outcome. Comparing GT with MIS tests whether task gains depend on correct observation content. Moreover, we introduce the coin-flip reward condition (COIN), which uses rewards independent of both the environment and the response to test whether gains remain without environment information in the reward. Figure~\ref{fig:audit_to_application} summarizes these four conditions.

Our results show that higher prediction accuracy does not necessarily imply better task performance. Training with mismatched targets reduces prediction accuracy significantly, but preserves substantial task gains over no-training baseline. Even training with independent random rewards expands pass@64. Compared with BASE (non-training baseline), trained agents more often weigh alternative actions and revise their choices during reasoning, while spending fewer interaction steps in repeated observation–action cycles.

These findings also suggest a simpler training design for multimodal agents. Training with true next observations in multimodal environments often requires an additional VLM to process screenshot outcomes into textual targets \citep{yu2025dynamind,ui_oceanus}. Since coverage expansion does not require environment information in the reward, we retain the post-training pipeline on VisualWebArena \citep{vwa} without constructing an observation-matching reward or adding a multimodal teacher for reward construction. Random-reward training still raises pass@64 by 14.3\% relative to BASE, consistent with the pattern observed in the text environments.

Our contributions are threefold:
\begin{itemize}[leftmargin=*,labelindent=0pt]
  \item We identify an overlooked issue in world-model post-training. Task gains are attributed to improved prediction ability, without accounting for other effects the training process itself may introduce.
  \item We design a controlled evaluation framework using mismatched targets and random rewards as controls. It measures prediction accuracy alongside task performance to assess the contribution of correct environment information.
  \item We conduct experiments on three benchmarks, showing that task gains from world-model post-training cannot be attributed solely to improved environment prediction. We further extend training to VisualWebArena with random rewards, which improves pass@64 by 14.3\% relative to BASE.
\end{itemize}

\section{Attribution Design}
\label{sec:framework}

\subsection{Methods under study}

We study two recent forms of world-model post-training. The RL-based method takes the interaction history, current observation, and executed action as input, and generates a reasoning trace followed by a next-observation prediction. It compares the predicted and observed next observations in an embedding space and optimizes the policy with GRPO \citep{rwml,grpo}. The distillation-based method supervises the same prediction through a teacher's token distribution, following the on-policy self-distillation (OPSD) procedure in COMAP \citep{comap}. The teacher additionally receives the true or mismatched next observation. We analyze representative implementations of these two methods.

\subsection{Experimental design}

The experimental design tests whether task gains require correct observation content. The RL-based experiments use the four core conditions in Table~\ref{tab:conditions}. GT, MIS, and COIN share the same training prompt, base model, optimizer, KL regularization, sampling temperature, and number of training steps. Only the reward construction differs. The distillation comparison holds its training pipeline fixed and compares GT with MIS by changing only the next-observation supervision. This comparison tests whether the effect of observation content on task performance is specific to GRPO.

\begin{wraptable}{r}{0.50\textwidth}
  \vspace{-8mm}
  \setlength{\belowcaptionskip}{10pt}
  \caption{Core RL conditions and their control roles.}
  \label{tab:conditions}
  \begin{center}
  \setlength{\tabcolsep}{3pt}
  \resizebox{\linewidth}{!}{%
    \begin{tabular}{@{}llll@{}}
      \toprule
      Condition & Target & Reward & Purpose \\
      \midrule
      BASE & none         & none                 & untrained reference \\
      GT   & ground truth & embedding similarity & RWML method \\
      MIS  & mismatched   & embedding similarity & content placebo \\
      COIN & unused       & Bernoulli($0.5$)     & zero-information control \\
      \bottomrule
    \end{tabular}%
  }
  \end{center}
  \vspace{-4mm}
\end{wraptable}

BASE provides an untrained reference. GT uses the next observation returned by the environment. MIS replaces it with another ground-truth observation from the same training distribution, preserving the prediction task and scoring procedure while changing the environmental content learned by the model. GT and MIS score each prediction against their respective targets using Qwen3-Embedding cosine similarity. Their comparison tests whether the task gain depends on correct observation content. COIN does not use an observation target, and its reward is independent of both the environment and the response. It measures the training effect that remains when the reward carries no environment information. We also test whether the link between a prediction and its score affects task gains. GT-PERM and MIS-PERM shuffle rewards within each group while preserving the group's scores, with results reported in Appendix~\ref{app:additional_results}.

We evaluate next-observation prediction accuracy and task performance. Prediction accuracy checks that the content intervention changes the model's next-observation predictions. For $N$ transitions that do not appear in the training trajectories, let $g_i$ be the task description, $o_i$ the current observation, $a_i$ the executed action, $o'_i$ the next observation returned by the environment, and $\hat{o}'_i$ the model prediction. An independent LLM judge receives all five fields and returns a binary verdict. The judge assesses whether the prediction correctly describes how the action changes the environment, and prediction accuracy averages these verdicts.
\begingroup
\refstepcounter{equation}\edef\judgeequationnumber{\theequation}
\refstepcounter{equation}\edef\predequationnumber{\theequation}
\begin{equation*}
J_i = J\!\left(g_i,o_i,a_i,o'_i,\hat{o}'_i\right) \in \{0,1\}
\quad (\judgeequationnumber)
\qquad
\operatorname{PredAcc} = \frac{100}{N}\sum_{i=1}^{N} J_i
\quad (\predequationnumber)
\end{equation*}
\endgroup

We measure task performance through \emph{selection} and \emph{coverage}. Selection reflects how reliably a single sampled trajectory completes the task, and we measure it with pass@1. Coverage reflects the range of tasks the agent can solve given repeated attempts, and we measure it with pass@64. For each task, we sample $n$ independent trajectories and use the unbiased pass@$k$ estimator
\begin{equation}
\widehat{\operatorname{pass@}k}
= 1 - \frac{\binom{n-c}{k}}{\binom{n}{k}},
\end{equation}
where $c$ of the $n$ trajectories succeed \citep{chen2021codex}. Pass@64 is the percentage of tasks solved at least once within 64 attempts.

To analyze how task gains manifest in action generation, we also compare the final actions selected by different conditions at the same decision points. With full chain-of-thought (CoT) generation, we compute overlap among the final actions. When CoT is skipped, we compare the log probabilities assigned to candidate actions. CoT analysis covers output length, references to observed entities, action revision, and generation of multiple candidates. The latter two assess whether the model broadens its search over candidate actions. We measure trajectory looping with TIDE's loop ratio \citep{tide}, the fraction of interaction steps spent repeating an observation--action cycle.

\subsection{Experimental setup}

The main experiments use two interactive text environments. ALFWorld \citep{alfworld} contains household tasks built on ALFRED and provides text observations. ScienceWorld \citep{sciworld} evaluates scientific reasoning through interaction with a text-simulated world. We use Qwen2.5-7B-Instruct \citep{qwen25} in both environments. The distillation comparison is conducted on ALFWorld.

We evaluate agents with and without supplied action candidates. In the \emph{list regime}, the prompt supplies the admissible-action list. In the \emph{no-list regime}, the prompt specifies the action format but omits the candidate set. ALFWorld supports both regimes, and its list regime matches the original RWML evaluation \citep{rwml}. ScienceWorld does not expose an admissible-action list and is therefore evaluated only in the no-list regime.

Prediction accuracy is computed on $N=1{,}200$ transitions absent from the training trajectories. We use DeepSeek-V4-Flash with temperature zero as the judge. For task evaluation, we sample 64 independent trajectories per task. Coverage comparisons pair conditions by task and use the two-sided exact McNemar test \citep{mcnemar1947note}.
\FloatBarrier

\section{Analysis Results}
\label{sec:results}

\subsection{Prediction accuracy and task performance do not move together}
\label{sec:dissociation}

Table~\ref{tab:main_results} summarizes the main prediction, task, and action-level results in the two text environments. Action failure counts interaction steps without an executed action, including parser, grammar, and state-precondition failures; average turns include all interaction steps.

The GT--MIS comparison tests whether task gains depend on correct observation content. On ALFWorld with Qwen2.5-7B-Instruct, replacing the ground-truth next observation with a mismatched one lowers prediction accuracy by 29.8\%. GT exceeds MIS by only 2.75\% in list-regime pass@1, while pass@64 barely changes. The learned next-observation content changes substantially, while the number of tasks solved within 64 attempts remains similar.

The same separation appears in ScienceWorld and privileged OPSD. Across the three settings, replacing GT with MIS lowers prediction accuracy by 15.3--61.6\%, while the absolute pass@1 difference is 0.51--2.75\% and the pass@64 difference is at most 3.0\%. ScienceWorld extends the result to a second environment, and privileged OPSD extends it to token-level self-distillation.

\begin{table}[t]
  \setlength{\belowcaptionskip}{6pt}
  \caption{Task-type pass@1 and overall results by evaluation regime and training method. AF is action failure rate and LR is loop ratio. Turns is the mean number of interaction steps per episode. RWML reports three-run ALFWorld success of $13.0 \pm 1.3\%$ for ReAct and $32.6 \pm 2.1\%$ for GRPO-GT using Qwen2.5-7B-Instruct \citep{rwml}.}
  \label{tab:main_results}
  \begin{center}
  \scriptsize
  \setlength{\tabcolsep}{2.4pt}
  \renewcommand{\arraystretch}{1.06}

  \newlength{\taskdomainwidth}
  \newlength{\awdomaincolwidth}
  \newlength{\swdomaincolwidth}
  \setlength{\taskdomainwidth}{0.38\linewidth}
  \setlength{\awdomaincolwidth}{\dimexpr(\taskdomainwidth-10\tabcolsep)/6\relax}
  \setlength{\swdomaincolwidth}{\dimexpr(\taskdomainwidth-8\tabcolsep)/5\relax}

  \newcommand{\awdomainhead}{%
    \begin{tabular}{@{}*{6}{>{\centering\arraybackslash}p{\awdomaincolwidth}}@{}}
      Pick & Clean & Heat & Cool & Look & Pick2
    \end{tabular}}
  \newcommand{\awdomains}[6]{%
    \begin{tabular}{@{}*{6}{>{\centering\arraybackslash}p{\awdomaincolwidth}}@{}}
      #1 & #2 & #3 & #4 & #5 & #6
    \end{tabular}}
  \newcommand{\swdomainhead}{%
    \begin{tabular}{@{}*{5}{>{\centering\arraybackslash}p{\swdomaincolwidth}}@{}}
      Meas. & Elec. & Class. & Chem. & Biol.
    \end{tabular}}
  \newcommand{\swdomains}[5]{%
    \begin{tabular}{@{}*{5}{>{\centering\arraybackslash}p{\swdomaincolwidth}}@{}}
      #1 & #2 & #3 & #4 & #5
    \end{tabular}}

  \begin{tabular*}{\linewidth}{@{\extracolsep{\fill}}lc@{\hspace{5pt}}cccccc@{}}
    \specialrule{\heavyrulewidth}{0pt}{0pt}
    \rowcolor[HTML]{E1E1E1}
    \multicolumn{8}{c}{\textbf{\emph{ALFWorld}}} \\
    \specialrule{\lightrulewidth}{0pt}{\belowrulesep}
    \multirow{2}{*}{\textbf{Condition}}
      & \textbf{Domain score: pass@1 (\%) $\uparrow$}
      & \multirow{2}{*}{\textbf{pass@1 (\%)}}
      & \multirow{2}{*}{\textbf{pass@64 (\%)}}
      & \multirow{2}{*}{\textbf{Pred. (\%)}}
      & \multirow{2}{*}{\textbf{AF (\%)}}
      & \multirow{2}{*}{\textbf{LR (\%)}}
      & \multirow{2}{*}{\textbf{Turns}} \\
    \cmidrule(lr){2-2}
      & \awdomainhead & & & & & & \\
    \midrule
    \multicolumn{8}{@{}l}{\textbf{\emph{List regime}}} \\
    BASE      & \awdomains{30.48}{4.93}{6.89}{5.64}{17.04}{0.15} & 11.48 & 56.20 & 45.75 & 63.28 & 24.69 & 28.96 \\
    GRPO-GT   & \awdomains{70.90}{23.14}{32.13}{30.26}{58.77}{5.56} & 37.30 & 83.94 & 57.00 & 19.35 & 3.26 & 23.41 \\
    GRPO-MIS  & \awdomains{63.45}{29.15}{31.37}{22.62}{50.76}{4.73} & 34.55 & 83.21 & 27.25 & 14.27 & 2.13 & 24.38 \\
    GRPO-COIN & \awdomains{58.45}{22.84}{26.28}{19.36}{41.63}{8.61} & 30.41 & 87.23 & 52.75 & 19.50 & 1.92 & 25.06 \\
    \addlinespace[1pt]
    OPSD-GT   & \awdomains{63.69}{21.58}{18.15}{19.84}{41.48}{2.67} & 29.29 & 72.99 & 59.10 & 37.57 & 5.79 & 25.99 \\
    OPSD-MIS  & \awdomains{58.29}{19.32}{15.95}{16.58}{42.39}{1.33} & 26.69 & 75.55 & 43.80 & 35.83 & 4.27 & 26.60 \\
    \midrule
    \multicolumn{8}{@{}l}{\textbf{\emph{No-list regime}}} \\
    BASE      & \awdomains{0.00}{0.00}{0.00}{0.00}{0.15}{0.00} & 0.02 & 1.09 & 45.75 & 98.38 & 66.30 & 30.00 \\
    GRPO-GT   & \awdomains{16.39}{5.55}{8.33}{3.53}{21.52}{0.23} & 8.95 & 52.92 & 57.00 & 65.32 & 11.21 & 28.87 \\
    GRPO-MIS  & \awdomains{25.87}{10.18}{15.38}{9.34}{25.96}{0.65} & 14.52 & 70.80 & 27.25 & 49.61 & 6.10 & 27.98 \\
    GRPO-COIN & \awdomains{35.33}{9.73}{9.90}{5.54}{17.79}{2.17} & 14.34 & 74.09 & 52.75 & 55.73 & 4.66 & 27.88 \\
    \addlinespace[1pt]
    OPSD-GT   & \awdomains{9.61}{1.99}{2.24}{1.12}{15.47}{0.11} & 4.77 & 44.89 & 59.10 & 71.49 & 12.83 & 29.46 \\
    OPSD-MIS  & \awdomains{11.10}{1.54}{1.96}{0.75}{15.63}{0.08} & 4.90 & 40.15 & 43.80 & 63.53 & 10.53 & 29.46 \\
    \specialrule{\heavyrulewidth}{\aboverulesep}{0pt}
    \rowcolor[HTML]{E1E1E1}
    \multicolumn{8}{c}{\textbf{\emph{ScienceWorld}}} \\
    \specialrule{\lightrulewidth}{0pt}{\belowrulesep}
    \multirow{2}{*}{\textbf{Condition}}
      & \textbf{Domain score: pass@1 (\%) $\uparrow$}
      & \multirow{2}{*}{\textbf{pass@1 (\%)}}
      & \multirow{2}{*}{\textbf{pass@64 (\%)}}
      & \multirow{2}{*}{\textbf{Pred. (\%)}}
      & \multirow{2}{*}{\textbf{AF (\%)}}
      & \multirow{2}{*}{\textbf{LR (\%)}}
      & \multirow{2}{*}{\textbf{Turns}} \\
    \cmidrule(lr){2-2}
      & \swdomainhead & & & & & & \\
    \midrule
    \multicolumn{8}{@{}l}{\textbf{\emph{No-list regime}}} \\
    BASE      & \swdomains{0.09}{0.09}{1.07}{0.00}{5.83} & 0.73 & 9.50 & 36.67 & 85.63 & 48.19 & 27.08 \\
    GRPO-GT   & \swdomains{0.00}{0.08}{2.24}{0.31}{14.90} & 1.65 & 11.50 & 71.67 & 75.85 & 38.94 & 25.77 \\
    GRPO-MIS  & \swdomains{0.09}{0.77}{1.92}{0.31}{4.90} & 1.14 & 14.50 & 10.08 & 71.24 & 45.44 & 27.26 \\
    GRPO-COIN & \swdomains{0.18}{0.23}{1.10}{1.25}{11.15} & 1.25 & 16.00 & 33.92 & 69.05 & 46.17 & 26.83 \\
    \bottomrule
  \end{tabular*}
  \end{center}
\end{table}

\subsection{Random reward also expands coverage}
\label{sec:decomposition}

Table~\ref{tab:main_results} shows that COIN expands coverage even when the reward contains no environment information. In the ALFWorld list regime, it raises pass@64 from 56.20\% for BASE to 87.23\%, reaching the coverage levels of GT and MIS. ScienceWorld also shows a coverage gain over BASE.

\begin{figure}[!t]
  \centering
  \includegraphics[width=\linewidth]{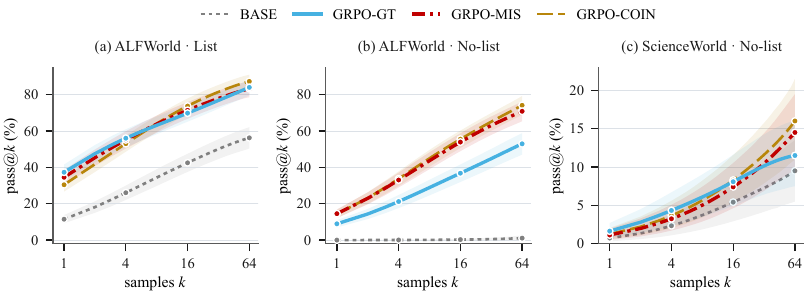}
  \caption{Pass@$k$ curves across training conditions on ALFWorld in the list (\textbf{left}) and no-list (\textbf{middle}) regimes, and ScienceWorld in the no-list regime (\textbf{right}). Shaded regions show task-level bootstrap 95\% intervals.}
  \label{fig:passk_curves}
\end{figure}

The pass@$k$ curves in Figure~\ref{fig:passk_curves} show that correct observation content primarily improves selection. In the ALFWorld list regime, GT reaches 37.30\% pass@1, compared with 30.41\% for COIN, but the two curves approach each other as $k$ increases. GT's advantage over MIS also disappears at $k=8$. The privileged OPSD comparison has the same shape, with the benefit of the correct target concentrated at pass@1. A reward constructed from the ground-truth next observation raises the probability that a single trajectory succeeds. COIN shows that coverage expansion does not require environment information in the reward.

\subsection{Correct content does not improve free-form generation}
\label{sec:free_form}

An admissible-action list confines decisions to candidates supplied by the environment. Removing the list requires the model to turn its reasoning into a complete action, giving observation content a more direct route to affect action generation. The results show the opposite. GT reaches 52.92\% pass@64 on ALFWorld, while COIN reaches 74.09\%. ScienceWorld shows the same ordering. Figure~\ref{fig:passk_curves} shows that GT solves the fewest tasks among the three core GRPO conditions in both environments.

More accurate next-observation prediction does not ensure that generated actions are executable in the current state. Table~\ref{tab:main_results} shows that GT has a higher action failure rate than MIS in the ALFWorld no-list regime, at 65.32\% versus 49.61\%.

\subsection{Training reshapes action generation without requiring correct observation content}
\label{sec:reasoning}

\begin{figure}[!t]
  \begin{center}
  \includegraphics[width=\linewidth]{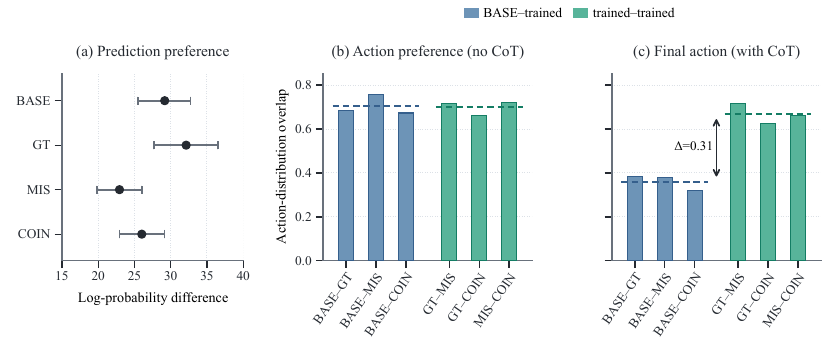}
  \end{center}
  \caption{Prediction preferences (\textbf{left}) and action-distribution overlap without CoT (\textbf{middle}) and with CoT (\textbf{right}). Error bars show 95\% bootstrap intervals. Dashed lines mark group means, and the arrow shows their difference.}
  \label{fig:reasoning_pathway}
\end{figure}

Figure~\ref{fig:reasoning_pathway} shows that GT and MIS differ in their predictive preferences while remaining close in their final-action distributions after reasoning. We compare both at 710 fixed ALFWorld decision points. Prediction preference measures how strongly the model favors the true next observation over a mismatched one, using the difference in their log probabilities. GT exceeds MIS by 9.19 nats, with a paired bootstrap 95\% interval of [7.65, 10.71]. Appendix~\ref{app:decision_probes} gives the scoring details.

We then examine action choice at these points in the list regime. Direct scoring assigns log probabilities to candidate actions without generating a CoT. We also let the model generate a complete CoT and measure its final action distribution. Under direct scoring, candidate-action preferences are similar across conditions. After a complete CoT, the final actions of GT, MIS, and COIN become more similar to one another and jointly diverge from BASE.

To characterize the changes in action generation, Table~\ref{tab:cot_profile} examines references to observed entities and comparisons between candidate actions, alongside output length. An action revision occurs when the model proposes an action and then abandons it for another. Multiple-candidate generation records whether it compares mutually exclusive actions within the same decision step. We use DeepSeek-V4-Flash to label these two behaviors on 300 uniformly sampled eligible steps per condition and regime, each with a parsed action and nonempty reasoning trace. The list regime supplies candidate actions, whereas the no-list regime requires the model to generate them.

\begin{figure}[!t]
  \centering
  \includegraphics[width=\linewidth]{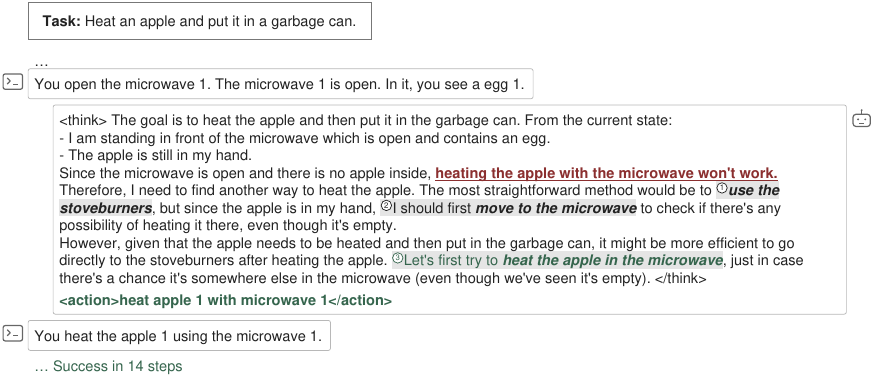}
  \begingroup
  \setlength{\fboxsep}{1pt}
  \caption{An MIS decision with action revision. \colorbox[HTML]{E5E5E5}{Gray Shading} highlights reasoning about candidate actions, \textcolor[HTML]{8B2F2F}{\underline{Red Underlining}} marks the incorrect prediction, and \textcolor[HTML]{35654D}{Green} marks the final correct action. Circled numbers indicate the order of candidate mentions.}
  \label{fig:prediction_revision_case_study}
  \endgroup
\end{figure}

\begin{table}[t]
  \setlength{\belowcaptionskip}{6pt}
  \caption{Chain-of-thought characteristics on ALFWorld. Tokens is the mean output length per step.}
  \label{tab:cot_profile}
  \begin{center}
  \scriptsize
  \setlength{\tabcolsep}{2.4pt}
  \renewcommand{\arraystretch}{1.06}
  \begin{tabular*}{\linewidth}{@{\extracolsep{\fill}}lcccccccc@{}}
    \toprule
    \multirow{2}{*}{\textbf{Condition}}
      & \multicolumn{4}{c}{\textbf{List}}
      & \multicolumn{4}{c}{\textbf{No-list}} \\
    \cmidrule(lr){2-5}\cmidrule(lr){6-9}
      & \textbf{Tokens}
      & \textbf{Entity (\%)}
      & \textbf{Revision (\%)}
      & \textbf{Candidates (\%)}
      & \textbf{Tokens}
      & \textbf{Entity (\%)}
      & \textbf{Revision (\%)}
      & \textbf{Candidates (\%)} \\
    \midrule
    BASE & 113 & 45.6 &  6.0 &  7.7 &  98 &  7.5 & 0.7 & 1.0 \\
    GRPO-GT   & 122 & 63.6 & 11.7 & 12.3 &  99 & 43.3 & 2.3 & 2.0 \\
    GRPO-MIS  & 132 & 57.7 &  9.3 &  9.7 & 112 & 40.2 & 4.0 & 3.7 \\
    GRPO-COIN & 150 & 53.0 & 11.7 & 12.7 & 131 & 27.9 & 5.7 & 8.3 \\
    \addlinespace[1pt]
    OPSD-GT   & 137 & 64.5 & 11.7 & 12.7 & 113 & 39.7 & 4.7 & 5.0 \\
    OPSD-MIS  & 133 & 58.3 & 10.3 & 11.3 & 108 & 42.1 & 3.7 & 4.3 \\
    \bottomrule
  \end{tabular*}
  \end{center}
\end{table}

The CoT measurements show shared behavioral changes across training methods and observation targets. In the no-list regime, entity-reference rates rise from 7.5\% for BASE to 27.9--43.3\% across the trained conditions. Action revision and multiple-candidate generation also increase for all five trained conditions in both regimes, extending the pattern from GRPO to OPSD. These changes appear under both mismatched supervision and random reward, so greater entity reference and broader action search do not require correct observation content. Providing an admissible-action list further increases revision and multiple-candidate rates for every condition.

Figure~\ref{fig:prediction_revision_case_study} shows MIS revising its action despite an incorrect prediction in its CoT. The prediction in red states that heating the apple in the microwave will not work. In the gray-shaded passages, it considers using the stoveburners and then moving to the microwave, before selecting the correct heating action in green. The environment confirms that the apple is heated, and the agent subsequently completes the task.

Broader action search also accompanies less trajectory looping. In the list regime, the loop ratio falls from 24.7\% for BASE to 1.9\% for COIN. The other GRPO conditions and the no-list regime show the same trend. All GRPO-trained conditions reduce looping, with COIN reaching the lowest loop ratio, so this behavioral change does not require correct observation content.

\section{From Attribution to Application}
\label{sec:application}

\subsection{Training design and evaluation setup}

An observation-matching reward requires predicted and observed outcomes to be comparable within the same representation. ALFWorld and ScienceWorld return text, so model predictions can be matched directly against the realized next observation. VisualWebArena instead returns screenshots to the browser agent \citep{vwa}. Continuing to use the same reward therefore requires an image-to-text model that converts the next screenshot into a textual target.

We compare BASE and COIN using Qwen2.5-VL-7B \citep{qwen25vl}. We retain the GRPO pipeline on VisualWebArena and replace the observation-matching reward with COIN, as shown in Figure~\ref{fig:vwa_application} (\textbf{left}). The agent still conditions its decisions on screenshot observations, while training uses neither an additional multimodal teacher nor a matched-observation target.

To prevent parameterized variants of the same intent template from entering both training and evaluation, we split tasks by template so that each template appears on only one side. The evaluation contains 201 tasks, with 64 trajectories sampled per task and condition. Alongside pass@1 and pass@64, we report the mean output tokens per step and mean interaction turns per episode.

\subsection{Testing the attribution finding on VisualWebArena}
\label{sec:vwa}

\begin{figure}[!t]
  \begin{center}
  \includegraphics[width=\linewidth]{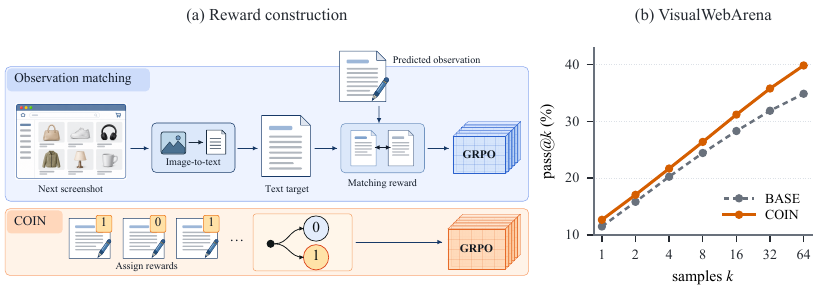}
  \end{center}
  \caption{VisualWebArena reward construction (\textbf{left}) and pass@$k$ for BASE and COIN (\textbf{right}).}
  \label{fig:vwa_application}
\end{figure}

\begin{table}[t]
  \setlength{\belowcaptionskip}{6pt}
  \caption{Task performance and interaction statistics on VisualWebArena.}
  \label{tab:vwa_results}
  \begin{center}
  \scriptsize
  \setlength{\tabcolsep}{2.4pt}
  \renewcommand{\arraystretch}{1.06}
  \begin{tabular*}{\linewidth}{@{\extracolsep{\fill}}llccccc@{}}
    \toprule
    \textbf{Website} & \textbf{Condition} & \textbf{pass@1 (\%)} & \textbf{pass@64 (\%)} & \textbf{LR (\%)} & \textbf{Tokens/step} & \textbf{Turns} \\
    \midrule
    \multirow{2}{*}{Classifieds} & BASE & 11.59 & 33.33 & 62.33 & 94.85 & 12.80 \\
     & COIN & 12.45 & 37.50 & 51.81 & 108.83 & 11.20 \\
    \midrule
    \multirow{2}{*}{Reddit} & BASE & 2.53 & 16.22 & 50.80 & 94.14 & 8.39 \\
     & COIN & 2.91 & 16.22 & 39.68 & 105.92 & 7.15 \\
    \midrule
    \multirow{2}{*}{Shopping} & BASE & 16.20 & 47.06 & 46.45 & 93.53 & 7.56 \\
     & COIN & 18.24 & 55.88 & 36.42 & 106.75 & 7.11 \\
    \midrule
    \multirow{2}{*}{\textbf{Overall}} & \textbf{BASE} & \textbf{11.48} & \textbf{34.83} & \textbf{56.86} & \textbf{94.41} & \textbf{10.22} \\
     & \textbf{COIN} & \textbf{12.66} & \textbf{39.80} & \textbf{46.14} & \textbf{107.85} & \textbf{9.07} \\
    \bottomrule
  \end{tabular*}
  \end{center}
\end{table}

If coverage expansion does not require environment information in the reward, training without matched-observation targets should also expand task coverage on VisualWebArena.

\begin{wraptable}{r}{0.50\textwidth}
  \vspace{-2mm}
  \setlength{\abovecaptionskip}{0pt}
  \setlength{\belowcaptionskip}{4pt}
  \caption{Episode termination types.}
  \label{tab:vwa_terminations}
  \begin{center}
  \scriptsize
  \setlength{\tabcolsep}{2.4pt}
  \renewcommand{\arraystretch}{1.06}
  \begin{tabular*}{\linewidth}{@{\extracolsep{\fill}}lcccc@{}}
    \toprule
    & \multicolumn{2}{c}{\textbf{BASE}} & \multicolumn{2}{c}{\textbf{COIN}} \\
    \cmidrule(lr){2-3}\cmidrule(lr){4-5}
    \textbf{Termination} & \textbf{Episodes} & \textbf{\%} & \textbf{Episodes} & \textbf{\%} \\
    \midrule
    Agent stop command & 5,444 & 42.32 & 6,822 & 53.03 \\
    Repeated-action limit & 5,583 & 43.40 & 4,703 & 36.56 \\
    Turn budget exhausted & 1,370 & 10.65 & 962 & 7.48 \\
    Consecutive parsing failures & 401 & 3.12 & 310 & 2.41 \\
    \bottomrule
  \end{tabular*}
  \end{center}
  \vspace{-3mm}
\end{wraptable}

The results in Table~\ref{tab:vwa_results} show that random-reward training expands task coverage. COIN improves pass@64 by 14.3\% relative to BASE. The coverage gains come from Shopping and Classifieds, where single-attempt success also improves. On Reddit, higher pass@1 accompanies unchanged pass@64, so successful attempts become more frequent without increasing the number of tasks solved. Figure~\ref{fig:vwa_application} (\textbf{right}) shows that COIN achieves higher pass@$k$ than BASE across the evaluated sampling budgets. The gap widens as $k$ increases, so repeated attempts reveal a larger coverage advantage over BASE.

Coverage expansion also accompanies shorter trajectories with less looping. Across all three websites, COIN produces more output tokens per step but uses fewer interaction turns per episode. The loop ratio falls on every website, decreasing from 56.86\% to 46.14\% overall. Table~\ref{tab:vwa_terminations} shows a corresponding change in how episodes end. Relative to BASE, COIN produces 16\% fewer trajectories terminated by the repeated-action limit and 30\% fewer that exhaust the step budget. Autonomous stops increase by 25\%. These changes echo the reduction in trajectory looping observed in the text environments.

The attribution findings help determine which training components to carry over to a visual environment. On VisualWebArena, they lead to a simpler reward design that still expands task coverage.

\section{Related Work}
\label{sec:related_work}

\paragraph{Agent tasks.} Agent benchmarks now span text-only long-horizon interaction, GUI and operating-system workflows, and coding and terminal environments. ALFWorld, ScienceWorld, and OdysseyArena evaluate multi-step reasoning and inductive discovery from textual feedback \citep{alfworld,sciworld,yan2026odysseyarena}. VisualWebArena, OSWorld, and OSWorld 2.0 add rendered interfaces and cross-application workflows \citep{vwa,xie2024osworld,yuan2026osworld2}, while SWE-bench and Terminal-Bench ground agents in code repositories and command-line environments \citep{jimenez2024swebench,merrill2026terminalbench}. Success across these tasks combines environment understanding, action generation, exploration, and recovery, so it cannot by itself identify which capability post-training changes.

\paragraph{World-model post-training for agents.} Recent methods incorporate environment prediction through reward agreement \citep{rwml,vagen}, auxiliary state or dynamics supervision \citep{envrl,tapo,paw}, and supervised, self-distilled, or jointly trained world models \citep{comap,spa_selfplay,describe_then_act,internalizing_future}. Beyond Next-Observation Prediction \citep{beyond_next_obs} asks the agent what it needs to know before acting. It then builds training targets from relevant state transitions. Some multi-turn RL methods instead aim to internalize environment dynamics without observation targets, by rewarding efficient interaction \citep{wmact} or task success with self-generated environment responses \citep{envace}. World-model post-training now reaches GUI and Android agents through UI-Oceanus and Dyna-Mind, as well as software-engineering and terminal agents through SWE-World and ECHO \citep{ui_oceanus,yu2025dynamind,sun2026sweworld,shrivastava2026echo}. A controlled study on Rubik's cubes \citep{better_wm_posttraining} finds that training to predict ground-truth cube states raises cube-state probe accuracy, and higher probe accuracy predicts larger gains from later GRPO training. In agent post-training, end-to-end gains still do not reveal whether improvement comes from correct environment content or from other training effects introduced by the prediction objective.

\paragraph{Uninformative rewards and content attribution.} RLVR can improve reasoning scores even when rewards contain little or no answer information \citep{spurious_rewards}, motivating placebo controls that match informative and uninformative prompts or targets \citep{form_not_content,kim_privileged_distill}. In OPSD for mathematical reasoning, a teacher that receives the solution to a different problem remains competitive with one that receives the paired reference solution \citep{ichihara_dissecting_opsd}. Conditioning the teacher on a reference solution pulls it toward that trajectory rather than toward correct solutions in general \citep{harne_privileged_biased}. Large-$k$ pass@$k$ further tests whether RL expands the solvable problem set \citep{yue_rlvr_boundary,rlvr_boundary_followup}. We extend these ideas to agent post-training. MIS changes learned environment content while preserving the prediction task and optimizer, whereas COIN removes environment information from the reward. Combining prediction accuracy with pass@1 and pass@64 separates content effects on single-attempt selection from changes in task coverage.

\section{Conclusion}
\label{sec:conclusion}

We audit the sources of task gains from world-model post-training using mismatched observation targets and independent random rewards. Across two interactive text environments and both RL and distillation, we observe the same separation: observation content changes prediction accuracy far more than task performance. Correct observation content primarily improves selection, while random-reward training also expands coverage. The action analysis further shows that this coverage expansion accompanies broader candidate-action search and less trajectory looping. Under free-form action generation, GT solves the fewest tasks among the three core GRPO conditions. This attribution also changes how training is designed for new environments. On VisualWebArena, COIN continues to expand coverage without an additional multimodal teacher or matched-observation target. World-model post-training gains can therefore be decomposed into content and optimization effects, and this decomposition directly identifies training dependencies that can be omitted in new environments.

\subsection*{AI use statement}

We used DeepSeek-V4-Flash and Qwen3-VL-32B to collect training trajectories, and Fable 5 and GPT-5.6-Sol to help polish the exposition of the paper.

\subsection*{Ethics statement}

This paper is a measurement audit of recent world-model post-training methods. It
runs entirely in simulated benchmark environments with publicly released base
models, and it involves no human subjects, no personal or private data, and no
deployment of the trained agents outside those benchmarks. Mismatched next
observations and uninformative rewards serve as experimental controls for
attribution.

\bibliography{iclr2027_conference}

@inproceedings{
alfworld,
title={{\{}ALFW{\}}orld: Aligning Text and Embodied Environments for Interactive Learning},
author={Mohit Shridhar and Xingdi Yuan and Marc-Alexandre Cote and Yonatan Bisk and Adam Trischler and Matthew Hausknecht},
booktitle={International Conference on Learning Representations},
year={2021},
url={https://openreview.net/forum?id=0IOX0YcCdTn}
}

@inproceedings{sciworld,
  title={Scienceworld: Is your agent smarter than a 5th grader?},
  author={Wang, Ruoyao and Jansen, Peter and C{\^o}t{\'e}, Marc-Alexandre and Ammanabrolu, Prithviraj},
  booktitle={Proceedings of the 2022 Conference on Empirical Methods in Natural Language Processing},
  pages={11279--11298},
  year={2022}
}

@inproceedings{vwa,
  title={Visualwebarena: Evaluating multimodal agents on realistic visual web tasks},
  author={Koh, Jing Yu and Lo, Robert and Jang, Lawrence and Duvvur, Vikram and Lim, Ming and Huang, Po-Yu and Neubig, Graham and Zhou, Shuyan and Salakhutdinov, Russ and Fried, Daniel},
  booktitle={Proceedings of the 62nd Annual Meeting of the Association for Computational Linguistics (Volume 1: Long Papers)},
  pages={881--905},
  year={2024}
}

@article{brown2020gpt3,
  title={Language models are few-shot learners},
  author={Brown, Tom and Mann, Benjamin and Ryder, Nick and Subbiah, Melanie and Kaplan, Jared D and Dhariwal, Prafulla and Neelakantan, Arvind and Shyam, Pranav and Sastry, Girish and Askell, Amanda and others},
  journal={Advances in neural information processing systems},
  volume={33},
  pages={1877--1901},
  year={2020}
}

@article{qwen3,
  title={Qwen3 technical report},
  author={Yang, An and Li, Anfeng and Yang, Baosong and Zhang, Beichen and Hui, Binyuan and Zheng, Bo and Yu, Bowen and Gao, Chang and Huang, Chengen and Lv, Chenxu and others},
  journal={arXiv preprint arXiv:2505.09388},
  year={2025}
}

@misc{qwen25,
      title={Qwen2.5 Technical Report}, 
      author={Qwen and : and An Yang and Baosong Yang and Beichen Zhang and Binyuan Hui and Bo Zheng and Bowen Yu and Chengyuan Li and Dayiheng Liu and Fei Huang and Haoran Wei and Huan Lin and Jian Yang and Jianhong Tu and Jianwei Zhang and Jianxin Yang and Jiaxi Yang and Jingren Zhou and Junyang Lin and Kai Dang and Keming Lu and Keqin Bao and Kexin Yang and Le Yu and Mei Li and Mingfeng Xue and Pei Zhang and Qin Zhu and Rui Men and Runji Lin and Tianhao Li and Tianyi Tang and Tingyu Xia and Xingzhang Ren and Xuancheng Ren and Yang Fan and Yang Su and Yichang Zhang and Yu Wan and Yuqiong Liu and Zeyu Cui and Zhenru Zhang and Zihan Qiu},
      year={2025},
      eprint={2412.15115},
      archivePrefix={arXiv},
      primaryClass={cs.CL},
      url={https://arxiv.org/abs/2412.15115}, 
}

@article{llama31,
  title={The llama 3 herd of models},
  author={Grattafiori, Aaron and Dubey, Abhimanyu and Jauhri, Abhinav and Pandey, Abhinav and Kadian, Abhishek and Al-Dahle, Ahmad and Letman, Aiesha and Mathur, Akhil and Schelten, Alan and Vaughan, Alex and others},
  journal={arXiv preprint arXiv:2407.21783},
  year={2024}
}

@article{grpo,
  title={Deepseekmath: Pushing the limits of mathematical reasoning in open language models},
  author={Shao, Zhihong and Wang, Peiyi and Zhu, Qihao and Xu, Runxin and Song, Junxiao and Bi, Xiao and Zhang, Haowei and Zhang, Mingchuan and Li, YK and Wu, Yang and others},
  journal={arXiv preprint arXiv:2402.03300},
  year={2024}
}

@article{chen2021codex,
  title={Evaluating large language models trained on code},
  author={Chen, Mark and Tworek, Jerry and Jun, Heewoo and Yuan, Qiming and Pinto, Henrique Ponde De Oliveira and Kaplan, Jared and Edwards, Harri and Burda, Yuri and Joseph, Nicholas and Brockman, Greg and others},
  journal={arXiv preprint arXiv:2107.03374},
  year={2021}
}

@article{mcnemar1947note,
  title={Note on the sampling error of the difference between correlated proportions or percentages},
  author={McNemar, Quinn},
  journal={Psychometrika},
  volume={12},
  number={2},
  pages={153--157},
  year={1947},
  publisher={Cambridge University Press \& Assessment}
}

@inproceedings{hafner2020dreamer,
  title = {Dream to Control: Learning Behaviors by Latent Imagination},
  author = {Hafner, Danijar and Lillicrap, Timothy and Ba, Jimmy and Norouzi, Mohammad},
  booktitle = {International Conference on Learning Representations (ICLR)},
  year = {2020},
  note = {arXiv:1912.01603}
}

@article{schrittwieser2020muzero,
  title={Mastering atari, go, chess and shogi by planning with a learned model},
  author={Schrittwieser, Julian and Antonoglou, Ioannis and Hubert, Thomas and Simonyan, Karen and Sifre, Laurent and Schmitt, Simon and Guez, Arthur and Lockhart, Edward and Hassabis, Demis and Graepel, Thore and others},
  journal={Nature},
  volume={588},
  number={7839},
  pages={604--609},
  year={2020},
  publisher={Nature Publishing Group UK London}
}

@inproceedings{hao2023rap,
  title={Reasoning with language model is planning with world model},
  author={Hao, Shibo and Gu, Yi and Ma, Haodi and Hong, Joshua and Wang, Zhen and Wang, Daisy and Hu, Zhiting},
  booktitle={Proceedings of the 2023 conference on empirical methods in natural language processing},
  pages={8154--8173},
  year={2023}
}

@article{vagen,
  title={Vagen: Reinforcing world model reasoning for multi-turn vlm agents},
  author={Wang, Kangrui and Zhang, Pingyue and Wang, Zihan and Gao, Yaning and Li, Linjie and Wang, Qineng and Chen, Hanyang and Lu, Yiping and Yang, Zhengyuan and Wang, Lijuan and others},
  journal={Advances in Neural Information Processing Systems},
  volume={38},
  pages={172871--172933},
  year={2025}
}

@article{envrl,
  title={EnvRL: Learn from Environment Dynamics in Agentic Reinforcement Learning},
  author={Wang, Zhitong and Li, Songze and Peng, Hao and Si, Shuzheng and Wang, Yi and Sun, Maosong and Li, Juanzi},
  journal={arXiv preprint arXiv:2606.17680},
  year={2026}
}

@article{tapo,
  title={TAPO: Transition-Aware Policy Optimization for LLM Agents},
  author={Li, Cong and Peng, Peixi and Zhao, Yisen and Hu, Xinyu and Liu, Shudong and Su, Zhan and Li, Zhuojian},
  journal={arXiv preprint arXiv:2607.27973},
  year={2026}
}

@article{paw,
  title={Policy and World Modeling Co-Training for Language Agents},
  author={Lu, Ning and Lin, Baijiong and Liu, Shengcai and Wu, Jiahao and Lv, Haoze and Wei, Yanbin and Zhu, Lingting and Qian, Shengju and Wang, Xin and Chen, Ying-Cong and others},
  journal={arXiv preprint arXiv:2606.02388},
  year={2026}
}

@article{wmact,
  title={Thinking by doing: Building efficient world model reasoning in llms via multi-turn interaction},
  author={Shu, Bao and Cai, Yan and Sun, Jianjian and Han, Chunrui and Yu, En and Zhao, Liang and Hu, Jingcheng and Zhang, Yinmin and Lv, Haoran and Peng, Yuang and others},
  journal={arXiv preprint arXiv:2511.23476},
  year={2025}
}

@article{envace,
  title={EnvACE: Internalizing Environment Dynamics via World Rehearsal for Agentic Reinforcement Learning},
  author={Xu, Zishan and Yao, Zhiyuan and Chen, Yuxin and Guo, Yifu and Lu, Zhengxi and Lu, Yuquan and Huang, Jinyang and Xu, Yan and Wang, Yasheng and Zhang, Weinan and others},
  journal={arXiv preprint arXiv:2608.06197},
  year={2026}
}

@article{internalizing_future,
  title={Internalizing the Future: A Unified Agentic Training Paradigm for World Model Planning},
  author={Zhang, Xuan and Zhou, Zhijian and Qiao, Lingfeng and Qin, Yulei and Li, Ke and Sun, Xing and Tan, Xiaoyu and Qu, Chao and Qi, Yuan},
  journal={arXiv preprint arXiv:2606.27483},
  year={2026}
}

@misc{spa_selfplay,
      title={Why Do LLM Agents Fail in Exploring New Environments? A World-Modeling Perspective}, 
      author={Shiqi Chen and Tongyao Zhu and Zian Wang and Jinghan Zhang and Kangrui Wang and Ruochen Zhou and Siyang Gao and Teng Xiao and Yee Whye Teh and Junxian He and Manling Li},
      year={2026},
      eprint={2510.15047},
      archivePrefix={arXiv},
      primaryClass={cs.LG},
      url={https://arxiv.org/abs/2510.15047}, 
}

@article{comap,
  title = {{COMAP}: Co-Evolving World Models and Agent Policies for {LLM} Agents},
  author = {Liu, Youwei and Wang, Jian and Wang, Hanlin and Li, Wenjie},
  journal = {arXiv preprint arXiv:2606.02372},
  year = {2026}
}

@article{ui_oceanus,
  title={UI-Oceanus: Scaling GUI agents with synthetic environmental dynamics},
  author={Wu, Mengzhou and Guo, Yuzhe and Cao, Yuan and Lu, Haochuan and Zhu, Songhe and Qu, Pingzhe and Chen, Xin and Qin, Kang and Wang, Zhongpu and Zhang, Xiaode and others},
  journal={arXiv preprint arXiv:2604.02345},
  year={2026}
}

@article{describe_then_act,
  title={Describe-Then-Act: Proactive Agent Steering via Distilled Language-Action World Models},
  author={Pappa, Massimiliano and Romani, Luca and Sacco, Valentino and Palma, Alessio and Lathuili{\`e}re, St{\'e}phane and Galasso, Fabio and Alameda-Pineda, Xavier and Spinelli, Indro},
  journal={arXiv preprint arXiv:2603.23149},
  year={2026}
}

@article{beyond_next_obs,
  title={Beyond Next-Observation Prediction: Agent-Authored World Modeling for Sequential Decision Making},
  author={Cai, Guangfeng and Yang, Kaibing and He, Shuo and Li, Yu and Yang, Shengtian and Lv, Jiaqi and Feng, Lei},
  journal={arXiv preprint arXiv:2606.25421},
  year={2026}
}

@article{better_wm_posttraining,
  title={Better world models can lead to better post-training performance},
  author={Gupta, Prakhar and Conklin, Henry and Leslie, Sarah-Jane and Lee, Andrew},
  journal={arXiv preprint arXiv:2512.03400},
  year={2025}
}

@inproceedings{spurious_rewards,
title={Spurious Rewards: Rethinking Training Signals in {RLVR}},
author={Rulin Shao and Shuyue Stella Li and Rui Xin and Scott Geng and Yiping Wang and Sewoong Oh and Simon Shaolei Du and Nathan Lambert and Sewon Min and Ranjay Krishna and Yulia Tsvetkov and Hannaneh Hajishirzi and Pang Wei Koh and Luke Zettlemoyer},
booktitle={Forty-third International Conference on Machine Learning},
year={2026},
url={https://openreview.net/forum?id=tqTNOpkP5j}
}

@article{yue_rlvr_boundary,
  title={Does reinforcement learning really incentivize reasoning capacity in llms beyond the base model?},
  author={Yue, Yang and Chen, Zhiqi and Lu, Rui and Zhao, Andrew and Wang, Zhaokai and Yue, Yang and Song, Shiji and Huang, Gao},
  journal={Advances in Neural Information Processing Systems},
  volume={38},
  pages={57654--57689},
  year={2025}
}

@article{rlvr_boundary_followup,
  title = {When {RLVR} Shrinks the Reasoning Boundary: Diagnosing Pass@$k$ Inversion},
  author = {Zhou, Todd},
  journal = {arXiv preprint arXiv:2607.20543},
  year = {2026}
}

@article{form_not_content,
  title={Form, Not Content? A Preregistered, Placebo-Controlled Evaluation of Learned Error-Conditioned Self-Repair Through Prompts and Weights in Frozen Small Code Models},
  author={Iscan, Mehmet},
  journal={arXiv preprint arXiv:2607.12962},
  year={2026}
}

@article{kim_privileged_distill,
  title={Why Does Self-Distillation (Sometimes) Degrade the Reasoning Capability of LLMs?},
  author={Kim, Jeonghye and Luo, Xufang and Kim, Minbeom and Lee, Sangmook and Kim, Dohyung and Jeon, Jiwon and Li, Dongsheng and Yang, Yuqing},
  journal={arXiv preprint arXiv:2603.24472},
  year={2026}
}

@article{ichihara_dissecting_opsd,
  title={Privileged Solutions or Context-Induced Teacher Behavior? Dissecting On-Policy Self-Distillation},
  author={Ichihara, Yuki and Iwase, Naoto and Quamar, Mohammad Atif and Komiyama, Junpei},
  journal={arXiv preprint arXiv:2608.09228},
  year={2026}
}

@article{harne_privileged_biased,
  title={Privileged, but biased: How pi-conditioned teachers break self-distillation},
  author={Harne, Sarthak and Karkar, Chinmay and Pandya, Yash and Awadallah, Ahmed and Nambi, Akshay},
  journal={arXiv preprint arXiv:2608.04794},
  year={2026}
}

@article{rwml,
  title={Reinforcement world model learning for LLM-based agents},
  author={Yu, Xiao and Peng, Baolin and Xu, Ruize and Shen, Yelong and He, Pengcheng and Nath, Suman and Singh, Nikhil and Gao, Jiangfeng and Yu, Zhou},
  journal={arXiv preprint arXiv:2602.05842},
  year={2026}
}

@article{patchworld,
  title={PatchWorld: Gradient-Free Optimization of Executable World Models for Agent Environments},
  author={Bai, Jiaxin and Guo, Yue and Dong, Yifei and Xiong, Jiaxuan and Zheng, Tianshi and Li, Yixia and Fang, Tianqing and Li, Yufei and Gao, Yisen and Huang, Haoyu and others},
  journal={arXiv preprint arXiv:2605.30880},
  year={2026}
}

@misc{qwen25vl,
      title={Qwen2.5-VL Technical Report}, 
      author={Shuai Bai and Keqin Chen and Xuejing Liu and Jialin Wang and Wenbin Ge and Sibo Song and Kai Dang and Peng Wang and Shijie Wang and Jun Tang and Humen Zhong and Yuanzhi Zhu and Mingkun Yang and Zhaohai Li and Jianqiang Wan and Pengfei Wang and Wei Ding and Zheren Fu and Yiheng Xu and Jiabo Ye and Xi Zhang and Tianbao Xie and Zesen Cheng and Hang Zhang and Zhibo Yang and Haiyang Xu and Junyang Lin},
      year={2025},
      eprint={2502.13923},
      archivePrefix={arXiv},
      primaryClass={cs.CV},
      url={https://arxiv.org/abs/2502.13923}, 
}

@article{yan2026odysseyarena,
  title={Odysseyarena: Benchmarking large language models for long-horizon, active and inductive interactions},
  author={Yan, Hang and Xu, Fangzhi and Sun, Qiushi and Wu, Jinyang and Huang, Zixian and Huang, Muye and Gong, Jingyang and Ding, Zichen and Cheng, Kanzhi and Wang, Yian and others},
  journal={arXiv preprint arXiv:2602.05843},
  year={2026}
}

@article{xie2024osworld,
  title={Osworld: Benchmarking multimodal agents for open-ended tasks in real computer environments},
  author={Xie, Tianbao and Zhang, Danyang and Chen, Jixuan and Li, Xiaochuan and Zhao, Siheng and Cao, Ruisheng and Hua, Toh J and Cheng, Zhoujun and Shin, Dongchan and Lei, Fangyu and others},
  journal={Advances in Neural Information Processing Systems},
  volume={37},
  pages={52040--52094},
  year={2024}
}

@article{yuan2026osworld2,
  title={OSWorld2. 0: Benchmarking Computer Use Agents on Long-Horizon Real-World Tasks},
  author={Yuan, Mengqi and Zhou, Zilong and Xiong, Xinzhuang and Wu, Weiming and Sun, Jiayang and Song, Jiamin and Cui, Kaiqian and Wang, Bowen and Wu, Haoyuan and Li, Yitong and others},
  journal={arXiv preprint arXiv:2606.29537},
  year={2026}
}

@inproceedings{jimenez2024swebench,
  title={Swe-bench: Can language models resolve real-world github issues?},
  author={Jimenez, Carlos E and Yang, John and Wettig, Alexander and Yao, Shunyu and Pei, Kexin and Press, Ofir and Narasimhan, Karthik},
  booktitle={International Conference on Learning Representations},
  volume={2024},
  pages={54107--54157},
  year={2024}
}

@inproceedings{merrill2026terminalbench,
  title={Terminal-bench: Benchmarking agents on hard, realistic tasks in command line interfaces},
  author={Merrill, Mike and Shaw, Alexander and Carlini, Nicholas and Li, Boxuan and Raj, Harsh and Bercovich, Ivan and Shi, Lin and Shin, Jeong and Walshe, Thomas and Buchanan, E Kelly and others},
  booktitle={International Conference on Learning Representations},
  volume={2026},
  pages={40903--40986},
  year={2026}
}

@inproceedings{yu2025dynamind,
  title={Dyna-mind: Learning to simulate from experience for better ai agents},
  author={Yu, Xiao and Peng, Baolin and Galley, Michel and Cheng, Hao and Wu, Qianhui and Kulkarni, Janardhan and Nath, Suman and Yu, Zhou and Gao, Jianfeng},
  booktitle={International Conference on Learning Representations},
  volume={2026},
  pages={32435--32459},
  year={2026}
}

@article{sun2026sweworld,
  title={Swe-world: Building software engineering agents in docker-free environments},
  author={Sun, Shuang and Song, Huatong and Huang, Lisheng and Jiang, Jinhao and Le, Ran and Lv, Zhihao and Chen, Zongchao and Hu, Yiwen and Luo, Wenyang and Zhao, Wayne Xin and others},
  journal={arXiv preprint arXiv:2602.03419},
  year={2026}
}

@article{shrivastava2026echo,
  title={Echo: Terminal agents learn world models for free},
  author={Shrivastava, Vaishnavi and Kauffmann, Piero and Awadallah, Ahmed and Papailiopoulos, Dimitris},
  journal={arXiv preprint arXiv:2605.24517},
  year={2026}
}

@article{tide,
  title={Tide: Trajectory-based diagnostic evaluation of test-time improvement in llm agents},
  author={Yan, Hang and Che, Xinyu and Xu, Fangzhi and Sun, Qiushi and Ding, Zichen and Cheng, Kanzhi and Zhang, Jian and Qin, Tao and Liu, Jun and Lin, Qika},
  journal={arXiv preprint arXiv:2602.02196},
  year={2026}
}
\bibliographystyle{iclr2027_conference}

\appendix
\clearpage

\section{Experimental Details}
\label{app:experimental_details}

\subsection{Data and Training Settings}
\label{app:training}

For ALFWorld, we use DeepSeek-V4-Flash to collect trajectories on the official training split. We then construct and preprocess next-observation prediction samples following the approach of RWML \citep{rwml}. For ScienceWorld, we use interaction trajectories from PatchWorld \citep{patchworld} and exclude all trajectories associated with the 200 AgentGym evaluation tasks. We randomly sample 20,000 transitions for each text environment using seed 42. For VisualWebArena, we collect training trajectories with Qwen3-VL-32B and construct transition samples after filtering invalid actions and removing repeated transitions. The task split is described in Appendix~\ref{app:vwa}.

ALFWorld and ScienceWorld use Qwen2.5-7B-Instruct, while VisualWebArena uses Qwen2.5-VL-7B-Instruct. Table~\ref{tab:training_settings} reports the optimization and rollout settings. Inputs exceeding the length limits are removed before training. We evaluate the final checkpoint from each run.

\begin{table}[!ht]
  \setlength{\belowcaptionskip}{6pt}
  \caption{Training hyperparameters. Batch and mini-batch sizes count prompts, rollout $n$ counts responses per prompt, and micro-batch size counts responses per GPU. Training samples are counted before length filtering.}
  \label{tab:training_settings}
  \centering
  \scriptsize
  \setlength{\tabcolsep}{2.4pt}
  \renewcommand{\arraystretch}{1.06}
  \begin{tabular*}{\linewidth}{@{\extracolsep{\fill}}lcccc@{}}
    \toprule
    & \multicolumn{3}{c}{\textbf{GRPO}} & \textbf{OPSD} \\
    \cmidrule(lr){2-4}\cmidrule(l){5-5}
    \textbf{Parameter} & \textbf{ALFWorld} & \textbf{ScienceWorld} & \textbf{VisualWebArena} & \textbf{ALFWorld} \\
    \midrule
    Optimizer & AdamW & AdamW & AdamW & AdamW \\
    Learning rate & $10^{-6}$ & $10^{-6}$ & $10^{-6}$ & $10^{-6}$ \\
    Adam $\beta_1,\beta_2$ & 0.9, 0.999 & 0.9, 0.999 & 0.9, 0.999 & 0.9, 0.999 \\
    Weight decay & 0.01 & 0.01 & 0.01 & 0.01 \\
    Learning rate schedule & Constant & Constant & Constant & Constant \\
    Warmup steps & 0 & 0 & 0 & 0 \\
    Batch size & 32 & 32 & 32 & 32 \\
    Mini-batch size & 32 & 32 & 32 & 32 \\
    Micro-batch size per GPU & 1 & 1 & 4 & 1 \\
    Rollout $n$ & 8 & 8 & 8 & 1 \\
    Temperature & 1.0 & 1.0 & 1.0 & 1.0 \\
    Top-$p$ & 1.0 & 1.0 & 1.0 & 1.0 \\
    Max. input length (tokens) & 1,536 & 2,560 & 8,192 & 1,536 \\
    Max. output length (tokens) & 512 & 512 & 512 & 512 \\
    Training samples & 20,000 & 20,000 & 4,478 & 20,000 \\
    Training epochs & 1 & 1 & 1 & 1 \\
    Training steps & 622 & 625 & 139 & 622 \\
    Update epochs per batch & 1 & 1 & 1 & 1 \\
    \midrule
    Policy clip $\epsilon$ & 0.2 & 0.2 & 0.2 & -- \\
    Dual-clip coefficient $c$ & 3.0 & 3.0 & 3.0 & -- \\
    Max. gradient norm & 1.0 & 1.0 & 1.0 & 1.0 \\
    KL coefficient & 0.01 & 0.01 & 0.01 & 0 \\
    Entropy coefficient & 0.001 & 0.001 & 0.001 & 0 \\
    JSD $\alpha$ & -- & -- & -- & 0.5 \\
    Importance sampling weight clip & -- & -- & -- & 2.0 \\
    \bottomrule
  \end{tabular*}
\end{table}

GT and MIS follow RWML's embedding-based binary reward \citep{rwml}. We encode the predicted and target observations with Qwen3-Embedding-8B. The reward is 1 when their cosine similarity exceeds 0.8, and 0 otherwise. MIS permutes target observations across training samples so that each sample receives a target different from its original, while leaving the inputs unchanged. COIN replaces this scoring procedure with an independent Bernoulli(0.5) reward. All GRPO conditions normalize advantages by the within-group reward standard deviation and use KL regularization toward the initial model. We freeze the visual encoder during VisualWebArena training.

In OPSD, both the student and teacher are initialized from Qwen2.5-7B-Instruct, and the teacher remains fixed throughout training. The teacher additionally receives the true next observation in GT or a mismatched observation in MIS. We minimize the generalized Jensen--Shannon divergence ($\alpha=0.5$) between the teacher and student distributions over the full vocabulary, with importance sampling weights clipped at 2.0. The loss is applied only to the student's generated next-observation span. We use no additional reference-model KL loss or entropy regularization.

\subsection{Evaluation Settings}
\label{app:evaluation}

Task evaluation uses temperature 1.0 and a maximum of 30 interaction steps per episode. The output limit per step is 1,024 tokens for ALFWorld and VisualWebArena, and 512 tokens for ScienceWorld. VisualWebArena additionally uses top-$p$ sampling with $p=0.9$. Task success is determined by each benchmark's evaluator.

For prediction evaluation, we uniformly sample 600 transitions from each of ALFWorld's in-domain and out-of-domain test pools. For ScienceWorld, we sample 1,200 transitions from the held-out AgentGym task pool. Sampling uses seed 42, and all conditions share the same samples within each environment. Next-observation predictions use greedy decoding with a 640-token output limit. The prediction and judge prompts are provided in Appendix~\ref{app:prompts}.

Confidence intervals for the pass@$k$ curves are computed by sampling tasks with replacement. Each sampled task retains all 64 rollouts, and the same task draws are used across conditions. We recompute the mean pass@$k$ for each draw and report the central 95\% of 10,000 bootstrap estimates. For the prediction-preference and action-overlap measurements, we instead resample the shared decision points 2,000 times, retaining each point's measurements across conditions.

\section{Additional Task Results and Reward-Pairing Diagnostics}
\label{app:additional_results}

Both GT and MIS score each prediction by how closely it matches an observation target. Replacing the target with a mismatched observation changes what is rewarded, but the reward still depends on the prediction. We test whether this dependence contributes to task gains.

For each training input, we first score the sampled next-observation predictions using the corresponding GT or MIS target. GT-PERM and MIS-PERM then randomly reassign these scores within the group before computing GRPO advantages. The scores are unchanged, but a prediction can receive a score computed for another prediction. All other training settings match the corresponding GT or MIS condition.

Table~\ref{tab:permutation_results} shows that shuffling rewards reduces both pass@1 and pass@64 in the two ALFWorld regimes. In the list regime, the pass@1 decrease is larger for GT than for MIS, at 8.87\% versus 2.82\%. The decreases are closer in the no-list regime. All four pass@64 decreases are significant under two-sided, task-paired McNemar tests, with Holm-adjusted $p<0.012$. Assigning each score to the prediction it evaluates therefore benefits task performance even with a mismatched target.

These task decreases do not track prediction accuracy. Shuffling rewards lowers GT's prediction accuracy by 5.00\% but raises MIS's by 22.33\%. MIS therefore provides another case in which better next-observation prediction accompanies worse task performance.

\begin{table}[ht]
  \setlength{\belowcaptionskip}{6pt}
  \caption{Task-type pass@1 and overall results before and after reward permutation on ALFWorld. $\Delta$ is PERM minus the original condition, computed before rounding. Rate changes are in \%, and Turns changes are steps per episode.}
  \label{tab:permutation_results}
  \centering
  \scriptsize
  \setlength{\tabcolsep}{2.4pt}
  \renewcommand{\arraystretch}{1.06}

  \newlength{\permdomainwidth}
  \newlength{\permcolwidth}
  \setlength{\permdomainwidth}{0.38\linewidth}
  \setlength{\permcolwidth}{\dimexpr(\permdomainwidth-10\tabcolsep)/6\relax}

  \newcommand{\permdomainhead}{%
    \begin{tabular}{@{}*{6}{>{\centering\arraybackslash}p{\permcolwidth}}@{}}
      Pick & Clean & Heat & Cool & Look & Pick2
    \end{tabular}}
  \newcommand{\permdomains}[6]{%
    \begin{tabular}{@{}*{6}{>{\centering\arraybackslash}p{\permcolwidth}}@{}}
      #1 & #2 & #3 & #4 & #5 & #6
    \end{tabular}}
  \begin{tabular*}{\linewidth}{@{\extracolsep{\fill}}lc@{\hspace{5pt}}cccccc@{}}
    \specialrule{\heavyrulewidth}{0pt}{0pt}
    \rowcolor[HTML]{E1E1E1}
    \multicolumn{8}{c}{\textbf{\emph{ALFWorld}}} \\
    \specialrule{\lightrulewidth}{0pt}{\belowrulesep}
    \multirow{2}{*}{\textbf{Condition}} & \textbf{Domain score: pass@1 (\%) $\uparrow$}
      & \multirow{2}{*}{\textbf{pass@1 (\%)}}
      & \multirow{2}{*}{\textbf{pass@64 (\%)}}
      & \multirow{2}{*}{\textbf{Pred. (\%)}}
      & \multirow{2}{*}{\textbf{AF (\%)}}
      & \multirow{2}{*}{\textbf{LR (\%)}}
      & \multirow{2}{*}{\textbf{Turns}} \\
    \cmidrule(lr){2-2}
      & \permdomainhead & & & & & & \\
    \midrule
    \multicolumn{8}{@{}l}{\textbf{\emph{List regime}}} \\
    GT & \permdomains{70.90}{23.14}{32.13}{30.26}{58.77}{5.56} & 37.30 & 83.94 & 57.00 & 19.35 & 3.26 & 23.41 \\
    GT-PERM & \permdomains{57.87}{21.36}{20.99}{19.57}{41.53}{3.16} & 28.43 & 71.53 & 52.00 & 23.25 & 4.90 & 24.86 \\
    $\Delta$ & \permdomains{-13.03}{-1.78}{-11.14}{-10.70}{-17.24}{-2.40} & -8.87 & -12.41 & -5.00 & +3.90 & +1.64 & +1.45 \\
    \addlinespace[2pt]
    MIS & \permdomains{63.45}{29.15}{31.37}{22.62}{50.76}{4.73} & 34.55 & 83.21 & 27.25 & 14.27 & 2.13 & 24.38 \\
    MIS-PERM & \permdomains{61.23}{22.39}{25.56}{21.43}{48.94}{6.86} & 31.72 & 75.55 & 49.58 & 21.91 & 3.09 & 24.67 \\
    $\Delta$ & \permdomains{-2.22}{-6.76}{-5.81}{-1.19}{-1.81}{+2.13} & -2.82 & -7.66 & +22.33 & +7.64 & +0.95 & +0.29 \\
    \addlinespace[2pt]
    \midrule
    \multicolumn{8}{@{}l}{\textbf{\emph{No-list regime}}} \\
    GT & \permdomains{16.39}{5.55}{8.33}{3.53}{21.52}{0.23} & 8.95 & 52.92 & 57.00 & 65.32 & 11.21 & 28.87 \\
    GT-PERM & \permdomains{16.08}{4.45}{1.80}{1.43}{11.19}{0.27} & 6.20 & 44.89 & 52.00 & 65.21 & 12.68 & 29.04 \\
    $\Delta$ & \permdomains{-0.32}{-1.10}{-6.53}{-2.11}{-10.33}{+0.04} & -2.75 & -8.03 & -5.00 & -0.11 & +1.47 & +0.17 \\
    \addlinespace[2pt]
    MIS & \permdomains{25.87}{10.18}{15.38}{9.34}{25.96}{0.65} & 14.52 & 70.80 & 27.25 & 49.61 & 6.10 & 27.98 \\
    MIS-PERM & \permdomains{22.59}{6.87}{9.78}{6.69}{19.56}{0.80} & 11.17 & 63.14 & 49.58 & 53.20 & 5.97 & 28.46 \\
    $\Delta$ & \permdomains{-3.28}{-3.31}{-5.61}{-2.65}{-6.40}{+0.15} & -3.35 & -7.66 & +22.33 & +3.59 & -0.12 & +0.49 \\
    \addlinespace[2pt]
    \bottomrule
  \end{tabular*}
\end{table}

\section{Action and Reasoning Analysis}
\label{app:reasoning}

\subsection{Prediction Preferences and Action Distributions}
\label{app:decision_probes}

The 710 shared decision points span early, middle, and late stages of sampled ALFWorld trajectories.

At each of the 710 decision points, the prediction probe holds the task, history, current observation, and executed action fixed. The mismatched next observation is also fixed across conditions. Prediction preference is the true next observation's log probability minus the mismatched observation's log probability, summing token log probabilities over each observation.

The prediction-preference probe in Figure~\ref{fig:reasoning_pathway} uses the ALFWorld prediction prompt in Appendix~\ref{app:prompts_training}, with \texttt{<next\_state>} prefilled before scoring each observation. Direct action scoring uses the ALFWorld list prompt in Appendix~\ref{app:prompts_execution}, with \texttt{<action>} prefilled before scoring each candidate action. Neither scoring procedure generates a reasoning trace.

For action generation with CoT, we sample 64 responses per condition at each decision point and estimate the final-action distribution from the frequency of each parsed action. Outputs without a parsed action are counted as a single failure category. For direct scoring without CoT, we sum token log probabilities over each candidate action and apply a softmax across the admissible candidates to obtain the action distribution.

For two conditions with action distributions $p$ and $q$ at the same decision point, we measure their overlap as
\begin{equation}
\operatorname{Overlap}(p,q)=\sum_a\min\{p(a),q(a)\},
\end{equation}
The sum runs over all actions in either distribution, assigning zero probability to an action absent from one of them. Overlap ranges from zero for disjoint distributions to one for identical distributions. Figure~\ref{fig:reasoning_pathway} reports the mean overlap across the shared decision points.

\subsection{Chain-of-Thought Measurements}
\label{app:cot_measurements}

Table~\ref{tab:cot_profile} characterizes CoT through output length, entity references, action revision, and comparison of candidate actions. Tokens is the mean number of tokens in the complete output at each step, including the reasoning text and final action. Entity is the percentage of steps whose reasoning text mentions a previously observed entity. We identify these references by matching object names and identifiers, such as \texttt{drawer 3}.

We use DeepSeek-V4-Flash to annotate action revision and candidate comparison. For each condition, we uniformly sample 300 steps from each of the list and no-list regimes, requiring a parsed action and nonempty reasoning text. The judge receives only the reasoning text and final action for the current step, without the training condition.

An action revision occurs when the agent proposes an action and then abandons it or chooses another. Candidate comparison occurs when the agent weighs two or more alternative actions for its next step. Sequential plans and searches across several locations each count as one candidate. For example, ``check the drawer, then check the cabinet'' is one plan. Both metrics report the percentage of sampled steps that meet the corresponding criterion.

In an independent review, 40 of 60 steps labeled as action revision are confirmed. False positives mainly arise when the judge treats the choice of where to start a sequential search as action revision.

\subsection{Additional Case Studies}
\label{app:case_studies}

\begin{figure}[!t]
  \centering
  \includegraphics[width=\linewidth]{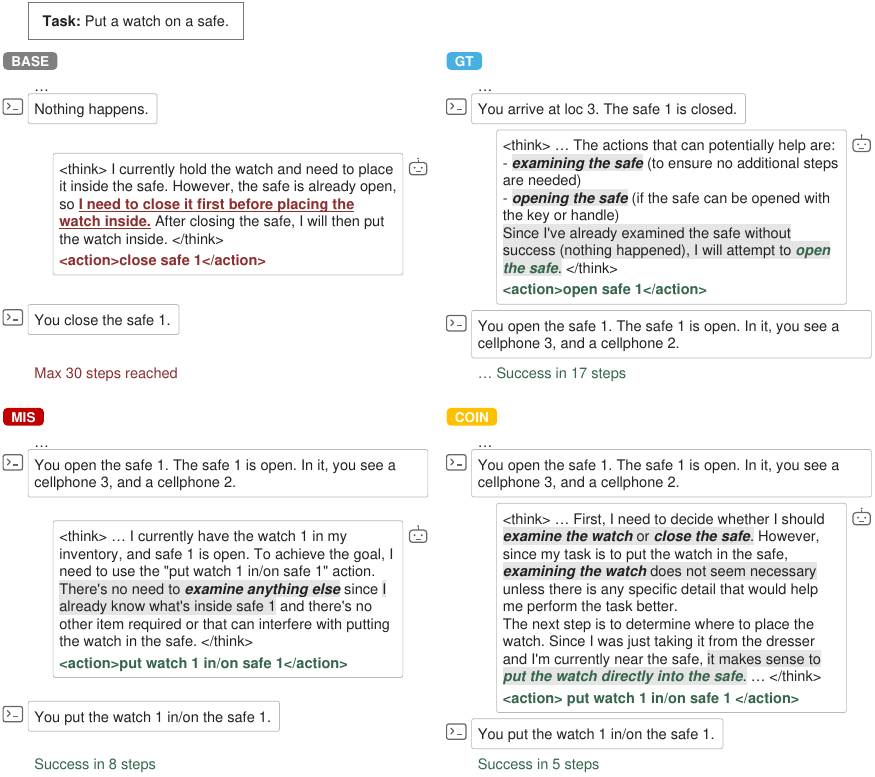}
  \caption{Trained agents compare or rule out candidate actions, including under mismatched supervision and random rewards.}
  \label{fig:watch_case_study}
\end{figure}

Figure~\ref{fig:watch_case_study} illustrates how candidate comparison guides an individual decision. MIS rules out further examination of the safe because its contents are already known. COIN considers examining the watch or closing the safe, then dismisses unnecessary examination and chooses to place the watch inside. Both complete the task.

\section{VisualWebArena Details}
\label{app:vwa}

\paragraph{Task split.}
We split tasks by intent template within each website, so no template appears in both training and evaluation. We train on 4,478 transition samples collected from rollouts on 437 tasks. For parallel evaluation, we exclude tasks such as placing orders or publishing posts because they change the shared website state and can affect other attempts' outcomes.\footnote{The \href{https://github.com/OpenGVLab/ScaleCUA/tree/main/evaluation/WebArenaLiteV2}{ScaleCUA WebArena-Lite-v2 evaluation guide} also notes interference between tasks sharing website instances.} This leaves 201 read-only tasks from the held-out set.

\paragraph{Paired results.}
A two-sided exact McNemar test gives $p=0.0414$ for the overall coverage difference. The 95\% bootstrap interval for the net gain is $[1,19]$ tasks, based on 2,000 resamples of task pairs.

\paragraph{Looping.}

LR in Table~\ref{tab:vwa_results} is the fraction of non-scroll steps spent repeating actions. After removing scroll commands, we identify adjacent identical action sequences and count the steps in the second occurrence. A sequence can contain one or more actions. Overlapping matches count each step once, and empty actions count as the same action. We pool repeated and total non-scroll step counts across all trajectories to compute LR.

\section{Prompt Templates}
\label{app:prompts}

The following templates are used for training, task execution, and evaluation. Fields in braces are filled with the corresponding task, observation, action, or history. Text templates are supplied as user messages before applying the model's chat template, except for the VisualWebArena system and user messages. Bracketed role and attachment headings indicate message structure and are not literal prompt text.

\subsection{Prediction Training}
\label{app:prompts_training}

The following complete templates are used for ALFWorld and ScienceWorld prediction training and prediction-accuracy evaluation. The history contains alternating observation and action entries from preceding steps, without reasoning traces.

\begin{promptbox}{Next-observation prediction: ALFWorld}
You are a world model for the ALFRED Embodied Environment. Given the interaction history, the current observation, and the action just taken, predict the resulting next observation.

Task: {task}

Interaction history:
{history}

Current observation: {current_observation}
Action taken: {action}

First reason step-by-step about the effect of the action within <think> </think> tags. Then give the predicted next observation within <next_state> </next_state> tags.
\end{promptbox}

\begin{promptbox}{Next-observation prediction: ScienceWorld}
You are a world model for the ScienceWorld text environment. Given the interaction history, the current observation, and the action just taken, predict the resulting next observation.

Task: {task}

Interaction history:
{history}

Current observation: {current_observation}
Action taken: {action}

First reason step-by-step about the effect of the action within <think> </think> tags. Then give the predicted next observation within <next_state> </next_state> tags.
\end{promptbox}

The complete OPSD student and teacher prompts are shown below. In the teacher prompt, \texttt{target\_next\_state} is the true next observation for OPSD-GT and the mismatched observation for OPSD-MIS.

\begin{promptbox}{OPSD student: prediction prompt}
You are a world model for the ALFRED Embodied Environment. Given the interaction history, the current observation, and the action just taken, predict the resulting next observation.

Task: {task}

Interaction history:
{history}

Current observation: {current_observation}
Action taken: {action}

First reason step-by-step about the effect of the action within <think> </think> tags. Then give the predicted next observation within <next_state> </next_state> tags.
\end{promptbox}

\begin{promptbox}{OPSD teacher: privileged prediction prompt}
You are a world model for the ALFRED Embodied Environment. Given the interaction history, the current observation, and the action just taken, predict the resulting next observation.

Task: {task}

Interaction history:
{history}

Current observation: {current_observation}
Action taken: {action}

First reason step-by-step about the effect of the action within <think> </think> tags. Then give the predicted next observation within <next_state> </next_state> tags.


[Privileged information — ground truth] The actual resulting next observation is:
{target_next_state}
\end{promptbox}

VisualWebArena prediction training uses the following template. The \texttt{<image>} marker is replaced by the current page screenshot with numbered element marks. Each history entry contains the preceding operation description and action command, formatted as \texttt{\{step\}.\ \{operation\} -> \{action\}}.

\begin{promptbox}{Next-page prediction: VisualWebArena}
You are a world model for a web browsing environment. Given the objective, the action history, the current page screenshot (with numbered element marks), and the action just taken, predict the resulting next page.

Objective: {objective}

Action history:
{history}

Current page: <image>
URL: {url}

Action taken: {action}

First reason step-by-step about the effect of the action within <think> </think> tags. Then describe the predicted next page within <next_state> </next_state> tags.
\end{promptbox}

\subsection{Task Execution}
\label{app:prompts_execution}

The ALFWorld list regime supplies the admissible actions at the current step. Here, \texttt{action\_history} contains alternating \texttt{Observation} and \texttt{Action} entries, numbered from one.

\begin{promptbox}{ALFWorld task execution: list regime}
You are an expert agent operating in the ALFRED Embodied Environment. Your task is to: {task_description}
Prior to this step, you have already taken {step_count} step(s). Below are the most recent {history_length} observations and the corresponding actions you took: {action_history}
You are now at step {current_step} and your current observation is: {current_observation}
Your admissible actions of the current situation are: [{admissible_actions}].

Now it's your turn to take an action.
You should first reason step-by-step about the current situation. This reasoning process MUST be enclosed within <think> </think> tags.
Once you've finished your reasoning, you should choose an admissible action for current step and present it within <action> </action> tags.
\end{promptbox}

The ALFWorld no-list regime uses the following complete prompt.

\begin{promptbox}{ALFWorld task execution: no-list regime}
You are an expert agent operating in the ALFRED Embodied Environment. Your task is to: {task_description}
Here are the actions you can take:
go to (receptacle): move to a receptacle
open (receptacle): open a receptacle
close (receptacle): close a receptacle
take (object) from (receptacle): take an object from a receptacle
put (object) in/on (receptacle): place an object in or on a receptacle
examine (something): examine a receptacle or an object
use (object): use an object
heat (object) with (receptacle): heat an object using a receptacle
clean (object) with (receptacle): clean an object using a receptacle
cool (object) with (receptacle): cool an object using a receptacle
slice (object) with (object): slice an object using a sharp object
Prior to this step, you have already taken {step_count} step(s). Below are the most recent {history_length} observations and the corresponding actions you took: {action_history}
You are now at step {current_step} and your current observation is: {current_observation}

Now it's your turn to take an action.
You should first reason step-by-step about the current situation. This reasoning process MUST be enclosed within <think> </think> tags.
Once you've finished your reasoning, you should choose an action for current step and present it within <action> </action> tags.
\end{promptbox}

ScienceWorld uses the following no-list template with its environment-specific action types.

\begin{promptbox}{ScienceWorld task execution: no-list regime}
You are an expert agent operating in the ScienceWorld text environment. Your task is to: {task_description}
Here are the types of actions you can take:
activate (object): turn on / activate an object
close (object): close a container or door
connect (object) to (object): connect one electrical terminal to another (e.g. to build a circuit)
deactivate (object): turn off / deactivate an object
disconnect (object): disconnect an electrical terminal
dunk (object) in (object): dunk an object into a container of liquid
eat (object): eat an object
flush (object): flush an object (e.g. toilet)
focus on (object): focus on the task-relevant object (needed to score on many tasks)
go (location): move to a connected location/room
inventory: list what you are carrying
look around: describe the current room
look at (object): examine an object
look in (object): look inside a container
mix (object): mix the contents of a container
move (object) to (object): move an object to a container/receptacle
open (object): open a container or door
pick up (object): pick up an object into your inventory
pour (object) in (object): pour a container's contents into another container
put down (object): drop the held object
read (object): read text on an object (e.g. sign, book)
task: print the current task description
use (object) on (object): use one object on another
wait: wait for up to 10 iterations
wait1: wait a single iteration
Prior to this step, you have already taken {step_count} step(s). Below are the most recent {history_length} observations and the corresponding actions you took: {action_history}
You are now at step {current_step} and your current observation is: {current_observation}

Now it's your turn to take an action.
You should first reason step-by-step about the current situation. This reasoning process MUST be enclosed within <think> </think> tags.
Once you've finished your reasoning, you should choose an action for current step and present it within <action> </action> tags.
\end{promptbox}

VisualWebArena uses the following system and user messages. The observation text contains only interactable element IDs and types, while element content is read from the screenshot. The response contains reasoning, a one-sentence operation description, and an action command. Each history entry is formatted as \texttt{\{step\}.\ \{operation\} -> \{action\}}.

\begin{promptbox}{VisualWebArena task execution}
[SYSTEM MESSAGE]
You are an autonomous intelligent agent tasked with navigating a web browser. You will be given web-based tasks. These tasks will be accomplished through the use of specific actions you can issue.

Here's the information you'll have:
The user's objective: This is the task you're trying to complete.
The observation: a text list of the interactable element ids and types on the current page, as [id] [tagType]. The text content of elements is NOT included — read it from the screenshot.
The current page screenshot: each interactable element is annotated with a numbered bounding box; the number matches the id in the observation.
The current web page's URL, and your action history so far.

The actions you can perform fall into several categories:

Page Operation Actions:
```click [id]```: This action clicks on an element with a specific id on the webpage.
```type [id] [content]```: Use this to type the content into the field with id. By default, the "Enter" key is pressed after typing unless press_enter_after is set to 0, i.e., ```type [id] [content] [0]```.
```hover [id]```: Hover over an element with id.
```press [key_comb]```: Simulates the pressing of a key combination on the keyboard (e.g., Ctrl+v).
```scroll [down]``` or ```scroll [up]```: Scroll the page up or down.

Tab Management Actions:
```new_tab```: Open a new, empty browser tab.
```tab_focus [tab_index]```: Switch the browser's focus to a specific tab using its index.
```close_tab```: Close the currently active tab.

URL Navigation Actions:
```goto [url]```: Navigate to a specific URL.
```go_back```: Navigate to the previously viewed page.
```go_forward```: Navigate to the next page (if a previous 'go_back' action was performed).

Completion Action:
```stop [answer]```: Issue this action when you believe the task is complete. If the objective is to find a text-based answer, provide the answer in the bracket. If the objective doesn't require an answer, use ```stop []```.

Your response must strictly follow this format (all three tags, in this order):
<think>Your step-by-step reasoning about the current state and what to do next.</think>
<operation>A one-sentence description of the operation you are about to perform.</operation>
<action>The action command, e.g. click [45]</action>

Rules:
1. Issue exactly one action per response, and it must be valid in the current page.
2. The action command must exactly follow the documented format. Any malformed output counts as a failure.
3. Issue the stop action when you believe the objective is achieved. Do not generate anything after the action tag.

[USER MESSAGE]
OBJECTIVE: {objective}

ACTION HISTORY:
{history}

OBSERVATION:
{obs_text}

URL: {url}

CURRENT PAGE: (see the screenshot below)

[IMAGE ATTACHMENTS]
{task_images_if_any}
{current_marked_screenshot}
\end{promptbox}

The user text is followed by any task-provided images and then the current marked screenshot as image attachments.

\subsection{Prediction-Accuracy Judging}
\label{app:prompts_prediction_judge}

The prediction-accuracy judge receives the task, current observation, executed action, true next observation, and predicted next observation in the following template.

\begin{promptbox}{Prediction-accuracy judge: ALFWorld}
You are an expert evaluator of world-model predictions for the ALFRED (ALFWorld) text environment.

Given the current observation and an action, an agent predicted the resulting next observation. You are given the GROUND-TRUTH next observation actually returned by the environment. Judge whether the agent's prediction is CORRECT about the EFFECT OF THE ACTION.

## Task the agent is doing
{task}

## Current observation
{current_obs}

## Action taken
{action}

## Ground-truth next observation (from the environment)
{ground_truth}

## Agent's predicted next observation
{prediction}

## How to judge
Judge whether the prediction captures the DYNAMICS — what this action does to the world state — matching the ground truth.

Rules:
1. IGNORE surface differences: wording, paraphrase, formatting, ordering, length, extra reasoning. The agent writes in its own words and does NOT copy the environment's phrasing. Judge meaning, not string overlap.
2. Judge the ACTION'S EFFECT: did it get right what the action does — arriving at a location, a receptacle being open/closed, picking up / putting an object, an action succeeding vs. failing ("Nothing happens."), a state change (heated/cooled/cleaned/sliced)?
3. Unpredictable specifics are NOT required: content only revealed by doing the action and not inferable beforehand — e.g. exactly which objects are on a surface you just walked to, or inside a container you just opened — does NOT need to match. Do not penalize different/guessed specific objects.
4. HALLUCINATION / OVER-PREDICTION is wrong: mark INCORRECT if the prediction adds events that did not happen (e.g. the action was only "go to X" but the prediction also opens it and lists contents), predicts success when the action failed (or vice versa), a wrong resulting location/state, or otherwise contradicts the ground-truth dynamics.

Verdict:
- CORRECT: dynamically consistent with the ground truth (right effect of the action), ignoring wording and unpredictable specifics, and without hallucinating events that did not occur.
- INCORRECT: wrong effect, contradicts the ground truth, or hallucinates/over-predicts.

First give a one-sentence justification. Then output the verdict on its own final line, exactly one of:
VERDICT: CORRECT
VERDICT: INCORRECT
\end{promptbox}

The complete ScienceWorld prediction-accuracy judge prompt is shown below.

\begin{promptbox}{Prediction-accuracy judge: ScienceWorld}
You are an expert evaluator of world-model predictions for the ScienceWorld text environment.

Given the current observation and an action, an agent predicted the resulting next observation. You are given the GROUND-TRUTH next observation actually returned by the environment. Judge whether the agent's prediction is CORRECT about the EFFECT OF THE ACTION.

## Task the agent is doing
{task}

## Current observation
{current_obs}

## Action taken
{action}

## Ground-truth next observation (from the environment)
{ground_truth}

## Agent's predicted next observation
{prediction}

## How to judge
Judge whether the prediction captures the DYNAMICS — what this action does to the world state — matching the ground truth.

Rules:
1. IGNORE surface differences: wording, paraphrase, formatting, ordering, length, extra reasoning. The agent writes in its own words and does NOT copy the environment's phrasing. Judge meaning, not string overlap.
2. Judge the ACTION'S EFFECT: did it get right what the action does — moving to a location, a door/container being opened/closed, picking up / moving an object, activating/deactivating a device, focusing on an object, an action succeeding vs. failing ("No known action matches that input." / "The door is not open."), or a physical state change (heated/cooled/melted/frozen/boiled/mixed)?
3. Unpredictable specifics are NOT required: content only revealed by doing the action and not inferable beforehand — e.g. exactly which objects are in a room you just entered, inside a container you just opened, or a measured numeric value (temperature/melting point) — does NOT need to match. Do not penalize different/guessed specific objects or numbers.
4. HALLUCINATION / OVER-PREDICTION is wrong: mark INCORRECT if the prediction adds events that did not happen (e.g. the action was only "go to X" but the prediction also opens a container and lists contents), predicts success when the action failed (or vice versa), a wrong resulting location/state, or otherwise contradicts the ground-truth dynamics.

Verdict:
- CORRECT: dynamically consistent with the ground truth (right effect of the action), ignoring wording and unpredictable specifics, and without hallucinating events that did not occur.
- INCORRECT: wrong effect, contradicts the ground truth, or hallucinates/over-predicts.

First give a one-sentence justification. Then output the verdict on its own final line, exactly one of:
VERDICT: CORRECT
VERDICT: INCORRECT
\end{promptbox}

\subsection{Chain-of-Thought Annotation}
\label{app:prompts_cot_judge}

The annotation prompt receives a single reasoning trace and its final action, without the training condition or task outcome. Table~\ref{tab:cot_profile} uses \texttt{rejected = yes} for action revision and \texttt{n\_candidates >= 2} for multiple-candidate generation. The complete prompt is reproduced below, including the additional fields collected during annotation.

\begin{promptbox}{Chain-of-thought annotation prompt}
You are annotating reasoning transcripts from an AI agent doing household tasks (navigating rooms; finding, moving, heating, cooling, cleaning objects; operating appliances). At each step the agent writes a private reasoning text (THINK) and then emits exactly one environment action (FINAL ACTION). You see one single step.

Label how the agent handled ACTION CANDIDATES inside THINK at this step.

Definitions:

1. Action candidate — a concrete action or plan the agent explicitly considers as its IMMEDIATE next move at this step.
   - A sequential plan ("go to the desk, then take the CD, then...") is ONE candidate: its first executable action together with its follow-ups. Later steps of a sequence are NOT separate candidates.
   - A sweep over several places/objects to check one after another ("I will examine the shelves, sidetables, and sofas", "I should check the sink basin and the cabinets") is ONE candidate (a sweep plan), NOT one per place. Enumerating possible locations and then choosing which to check FIRST, keeping the rest for later ("options are the countertops, drawers, or shelves... I will start with the countertops"), is STILL one sweep candidate — prioritizing within a sweep is not a rejection.
   - Conditional plans for future steps ("if it is not there, I will check the drawer") are NOT candidates at this step.
   - Investigative moves phrased as "check if/whether X", "see if X", "let's check whether..." ARE candidates.
   - A candidate ALSO counts when the agent raises an action and sets it aside, even without saying "I will": the goal action deemed not possible yet ("it is not possible to put the hot egg in the garbagecan without first...", so it does something else → the garbagecan action was a candidate, rejected), or the eliminated side of an "X or Y" consideration ("I need a microwave or a stove... no microwave is available, so the stove" → the microwave option was a candidate, rejected; "I can either examine the shelf again to confirm, or just take the item. Since the item was already seen there, I will take it." → the examine option was a candidate, rejected).
   - Count separate candidates ONLY when options are weighed against each other for the next move: "X or Y", "either X or Y", preferring one over another, or proposing X and then displacing it with Y.
   - A candidate that is proposed and then abandoned or displaced still counts as a candidate.
   - Restating the same action in different words is still the same ONE candidate.

2. rejected — "yes" iff the agent considered some candidate for this step and then dismissed or displaced it: explicitly ("but", "however", "instead", "won't work", "no need", "not possible yet") or by switching to a different choice after proposing one — including reordering ("check X first... actually, check Y first"), which displaces the X-first candidate. NOT rejection: deferring a later step of its own sequential plan; choosing where to start within a sweep (queued targets are deferred, not rejected); explaining why a PREVIOUS step failed.

3. rejection_basis — only when rejected="yes" (else "n.a."). If several candidates were rejected, label the rejection whose reasoning is most explicit.
   - "predicted-consequence": the stated reason simulates what WOULD happen if the action were taken — a hypothetical outcome or failure that is NOT just a restatement of something already observed. Examples: "the microwave is closed, so I cannot put the mug in yet"; "washing it in the sink would get the book wet"; "taking the knife first would leave no free hand for the plate". The prediction must come from general knowledge of how the world works, not from failures already observed.
   - "history-memory": the stated reason cites what already happened — past actions or past observations. Examples: "I already checked drawer 1"; "the last observation showed the desk has no CD, so examining it again is pointless"; "that action failed before"; "I already have the mug". If the rejection merely restates a past observation or past action (even phrased with "would" or wrapped in a conditional — "if it is not functioning, as suggested by the repeated 'Nothing happens' responses" → the evidence is past observations, so history-memory), it belongs HERE, not in predicted-consequence.
   - "other": any other basis — preference or ordering with no stated outcome, a mere possibility ("it might also be on the countertop"), unstated reason, task-rule citation, etc.

4. final_match — compare FINAL ACTION against the LAST SURVIVING candidate (the most recently declared one, after any displacements):
   - "last": the action executes that candidate — directly, as an explicit enabling first step toward it (candidate "put the pillow on the bed", action "go to bed 1" → "last"), as any step of that sequential plan even if it skips ahead (candidate "examine the coffeetable, then take the remote", action "take remotecontrol 1" → "last"; candidate "put the lamp aside and inspect the bowl", action "examine bowl 1" → "last"), or on one member of a named sweep set (candidate "examine the shelves and sidetables", action "examine shelf 1" → "last").
   - "earlier": the action executes a candidate declared EARLIER than the last surviving one — including a candidate that had been rejected or displaced. Example: THINK says "I should check the sinkbasins first. However, the soapbar might also be on the countertop, so I will check that first." and FINAL ACTION is "examine sinkbasin 1" → "earlier" (the agent reverted to the displaced sinkbasins-first option).
   - "none": it matches no declared candidate (or there are no candidates).

Output STRICT JSON as the LAST thing you write (no trailing text):
{"candidates": ["<short paraphrase of each candidate, in order of appearance>"], "n_candidates": <0|1|2|3>, "rejected": "<yes|no>", "rejection_basis": "<predicted-consequence|history-memory|other|n.a.>", "final_match": "<last|earlier|none>", "evidence": "<quote of up to 25 words supporting rejected/rejection_basis, or empty string>"}

where n_candidates is the number of distinct candidates capped at 3 (use 3 for "3 or more"); it must equal the length of "candidates" capped at 3. rejection_basis is "n.a." if and only if rejected is "no".

THINK:
{think}

FINAL ACTION: {action}
\end{promptbox}

\end{document}